# IndoorBEV: A Lightweight Real-Time LiDAR BEV Perception System for Indoor Mobile Robots

**Haichuan Li**
haichuan,li@utu.fi
University of Turku

## Abstract

Efficient indoor LiDAR perception is challenging because mobile robots must understand cluttered three-dimensional environments under strict latency and memory constraints. Existing point-based and voxel-based methods often incur substantial computational overhead, whereas conventional bird's-eye-view (BEV) representations improve efficiency at the cost of discarding vertical geometric information. We present IndoorBEV, a lightweight LiDAR perception framework that mitigates this tradeoff through a height-aware BEV representation and geometry-conditioned feature fusion. IndoorBEV summarizes the vertical point distribution in each BEV cell using statistical height features and multi-frequency height encoding, allowing informative three-dimensional cues to be processed efficiently by two-dimensional convolutions. A lightweight encoder then integrates complementary geometric features with multi-scale local representations and compact global scene context. Decoupled dense prediction heads jointly produce semantic BEV maps and oriented object bounding boxes. IndoorBEV contains only 0.6M parameters and requires 2.3 MB of model storage. On an NVIDIA AGX Orin, it uses 21.52 MB of GPU memory per inference and achieves a mean latency of 169.6 ms under a 200 ms perception deadline, with a deadline miss ratio of 1.8%. Evaluations on simulated scenes, real-world robot scans, and an open-source indoor point-cloud dataset demonstrate a favorable tradeoff among perception accuracy, latency, and memory consumption. These results indicate that explicitly encoding vertical geometry within a compact BEV representation provides an effective approach to resource-efficient indoor LiDAR perception.

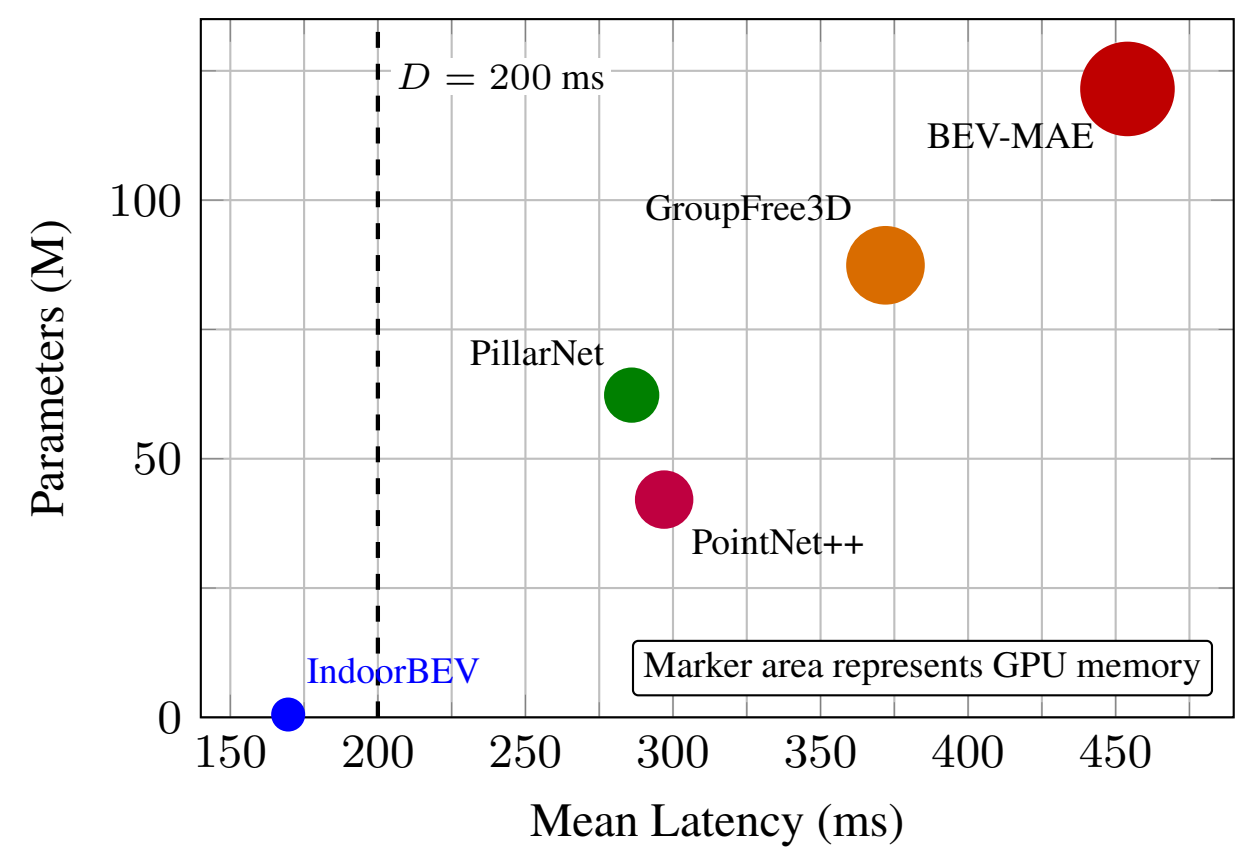

Figure 1: Model complexity, mean latency, and peak inference memory on the NVIDIA AGX Orin. Marker area represents peak GPU memory usage. IndoorBEV contains 0.54M parameters and achieves a mean latency of 169.6 ms, remaining below the 200 ms target deadline while using substantially fewer parameters and less inference memory than the evaluated baselines.

## Introduction

Autonomous mobile robots must continuously perceive obstacles, moving agents, and structural elements under tight computational and timing constraints. Unlike offline 3D scene understanding, onboard perception operates within a closed-loop pipeline in which excessive or highly variable latency reduces the time available for planning, collision avoidance, and control. Practical indoor perception therefore requires a favorable balance among prediction accuracy, latency, and memory consumption. LiDAR provides reliable geometric measurements for indoor perception, but directly processing its sparse and irregular point clouds is computationally expensive. Point-based and voxel-based methods preserve detailed 3D geometry, yet their neighborhood aggregation or volumetric processing often incurs substantial computational and memory overhead. Bird's-eye-view (BEV) representations offer a more efficient alternative by transforming point clouds into structured 2D feature maps. However, conventional BEV projection collapses the vertical dimension and may discard information needed to distinguish indoor objects with similar ground-plane footprints. Efficient indoor perception must therefore retain informative vertical cues without reintroducing expensive 3D processing.

We present IndoorBEV, a lightweight LiDAR perception framework that addresses this tradeoff through a height-aware BEV representation and axis-conditioned feature fusion. IndoorBEV encodes the local vertical point distribution using statistical features and multi-frequency height encoding, enabling efficient 2D processing while mitigating the information loss caused by BEV projection. A lightweight encoder combines complementary geometric transformations with multi-scale local features and compact global context. The resulting representation supports joint dense object-class prediction and oriented bounding-box regression. Indoor-BEV is further integrated into a deadline-aware perception pipeline for embedded mobile-robot deployment.

We evaluate IndoorBEV on simulated indoor scenes, real-world robot scans, and an open-source indoor point-cloud dataset. The evaluation considers both perception quality and embedded deployment efficiency, demonstrating the potential of height-aware BEV learning for resource-constrained indoor perception. The main contributions of this work are:

- We introduce a height-aware BEV representation that encodes cell-wise height distributions through statistical features and multi-frequency height encoding, mitigating the loss of vertical information caused by conventional BEV projection while retaining efficient 2D processing.
- We develop a lightweight axis-conditioned fusion network that combines complementary feature transformations, multi-scale local features, and compact global scene context to jointly predict dense object-class BEV maps and oriented object bounding boxes.
- We integrate IndoorBEV into an embedded mobile-robot perception pipeline and conduct a unified evaluation across simulated, real-world, and open-source indoor data, measuring perception accuracy together with model size, inference memory, latency distributions, and deadline compliance.

## Related Work

**Efficient LiDAR and BEV Perception.** Point-based methods, including PointNet (Qi et al. 2017a), PointNet++ (Qi et al. 2017b), DGCNN (Wang et al. 2019), KPConv (Thomas et al. 2019), RandLA-Net (Hu et al. 2020), and Point Transformer (Zhao et al. 2021), learn directly from irregular point sets and preserve detailed geometric information. Voxel-based and sparse-convolutional methods, such as VoxelNet (Zhou and Tuzel 2018), SECOND (Yan, Mao, and Li 2018), SparseConvNet (Graham, Engelcke, and van der Maaten 2018), MinkowskiNet (Choy, Gwak, and Savarese 2019), and Cylinder3D (Zhu et al. 2021), instead discretize 3D space to support structured feature extraction. Although effective, point-neighborhood aggregation and 3D feature processing can introduce substantial computational and memory overhead. BEV and pillar-based representations reduce this cost by transforming point clouds into structured 2D feature maps. Representative approaches include MV3D (Chen et al. 2017), PIXOR (Yang, Luo, and Urtasun 2018), PointPillars (Lang et al. 2019), CenterPoint (Yin, Zhou, and Krahenbuhl 2021), PillarNet (Shi et al. 2022), and BEV-MAE (Chen et al. 2023). Related BEV methods have also explored temporal reasoning, depth-aware projection, multi-modal fusion, and dense scene representations (Li et al. 2022b,a; Liu et al. 2023; Teng et al. 2024; Zhou et al. 2021; Cao and De Charette 2022). However, many existing methods target outdoor driving, depend on camera or multi-sensor inputs, or employ large backbones. IndoorBEV focuses on indoor perception and retains vertical geometric cues through a compact height-aware BEV representation.

**Indoor 3D Scene Understanding.** Datasets including NYUv2 (Silberman et al. 2012), SUN RGB-D (Song, Lichtenberg, and Xiao 2015), S3DIS (Armeni et al. 2016), ScanNet (Dai et al. 2017), Matterport3D (Chang et al. 2017), and LiDAR-Net (Guo et al. 2024) have supported research on indoor semantic segmentation and object detection. Detection methods such as VoteNet (Qi et al. 2019), ImVoteNet (Qi et al. 2020), H3DNet (Zhang et al. 2020), 3DETR (Misra et al. 2021), GroupFree3D (Liu et al. 2021), FCAF3D (Rukhovich, Vorontsova, and Konushin 2022), and CAGroup3D (Wang et al. 2022) improve localization through voting, geometric primitives, transformers, or fully convolutional prediction. These methods primarily emphasize perception accuracy, whereas IndoorBEV jointly considers semantic and geometric prediction together with the latency and memory constraints of embedded mobile robots.

**Real-Time Robotic Deployment.** Robotic perception operates within a closed-loop pipeline, making latency and resource consumption important in addition to prediction accuracy. Systems such as MakeWay (Xu et al. 2025a) and LV-DOT (Xu et al. 2025b) illustrate how semantic and geometric perception can support indoor navigation and dynamic-obstacle handling. ROS2 provides DDS communication, Quality-of-Service policies, and executor-based scheduling for modular robotic systems (Quigley et al. 2009; Maruyama, Kato, and Azumi 2016; Macenski et al. 2022), while prior studies have analyzed response time, callback scheduling, and end-to-end latency in ROS2 processing chains (Casini et al. 2019; Ye et al. 2023; Bédard, Dagenais et al. 2022). Complementary deployment techniques, including mixed precision, quantization, pruning, distillation, and optimized inference engines, can further reduce computational overhead (Micikevicius et al. 2017; Jacob et al. 2018; Han et al. 2015; Hinton, Vinyals, and Dean 2015; NVIDIA Corporation; Chen et al. 2018; ONNX Runtime Contributors). IndoorBEV complements these techniques through algorithm design tailored to resource-constrained perception.

## Methodology

### Overview

Given a LiDAR point cloud

$$P = \{p_i\}_{i=1}^{N}, \qquad p_i = (x_i, y_i, z_i, r_i),$$

where $(x_i, y_i, z_i)$ denotes the coordinate of point $i$ and $r_i$ denotes its intensity, IndoorBEV predicts a dense object-class BEV map and an oriented bounding-box regression map. As illustrated in Fig. 2, IndoorBEV consists of four stages. First, points within a predefined region of interest are converted into a fixed-resolution height-aware BEV representation. Second, multi-frequency height encoding augments the BEV features with vertical geometric cues. Third, a lightweight axis-conditioned encoder combines parallel feature transformations with multi-scale local features and compact global scene context. Finally, decoupled prediction heads produce dense object-class logits and oriented bounding-box parameters. The pipeline operates on fixed-size 2D feature maps and avoids dense 3D feature volumes, iterative proposal refinement, and point-neighborhood search during inference.

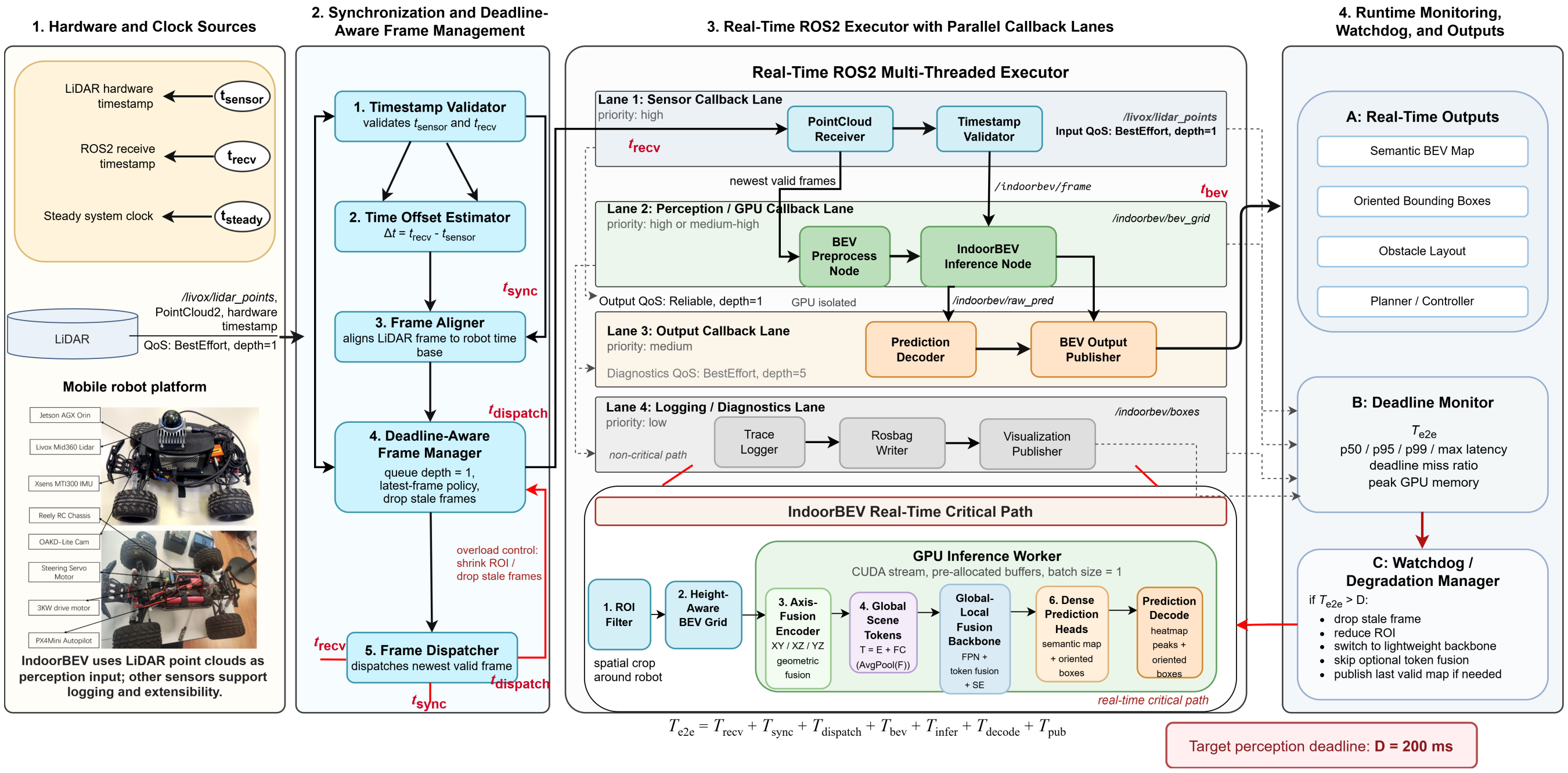

Figure 2: Deadline-aware ROS2 IndoorBEV real-time perception architecture. Hardware-timestamped LiDAR frames from the mobile robot are validated, time-aligned, and dispatched through a deadline-aware frame manager with a bounded latest-frame queue. The ROS2 executor separates sensor reception, GPU perception, output publication, and logging/diagnostics into parallel callback lanes to reduce interference between the real-time critical path and non-critical tasks. The IndoorBEV critical path performs ROI filtering, height-aware BEV construction, axis-fusion encoding, global scene token fusion, global-local feature fusion, dense prediction, and output decoding. Timing markers are collected by the deadline monitor to compute end-to-end latency, tail latency, deadline miss ratio, and peak GPU memory. When overload is detected, the watchdog can trigger degradation actions such as stale-frame dropping, ROI reduction, lightweight inference, or reuse of the last valid map.

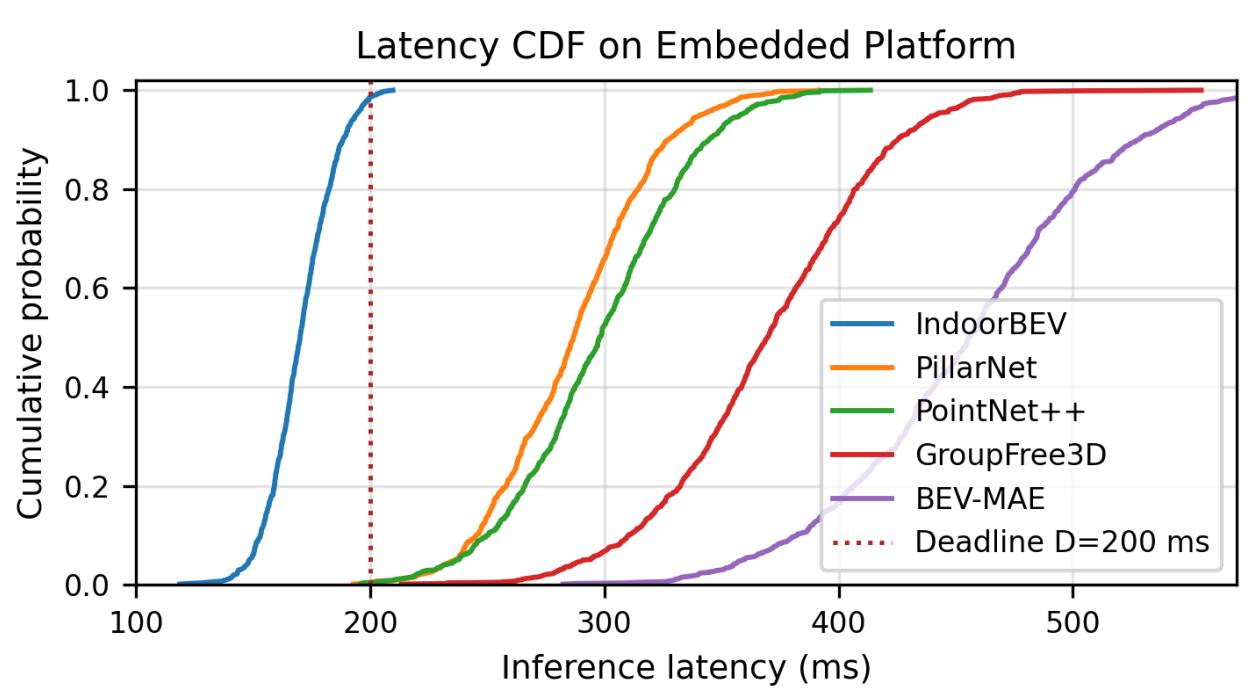

Figure 3: Cumulative distribution of end-to-end latency on the NVIDIA AGX Orin. The vertical dashed line marks the 200 ms perception deadline. IndoorBEV satisfies the deadline for 98.2% of the evaluated frames, whereas the adapted baselines exceed it for all measured frames.

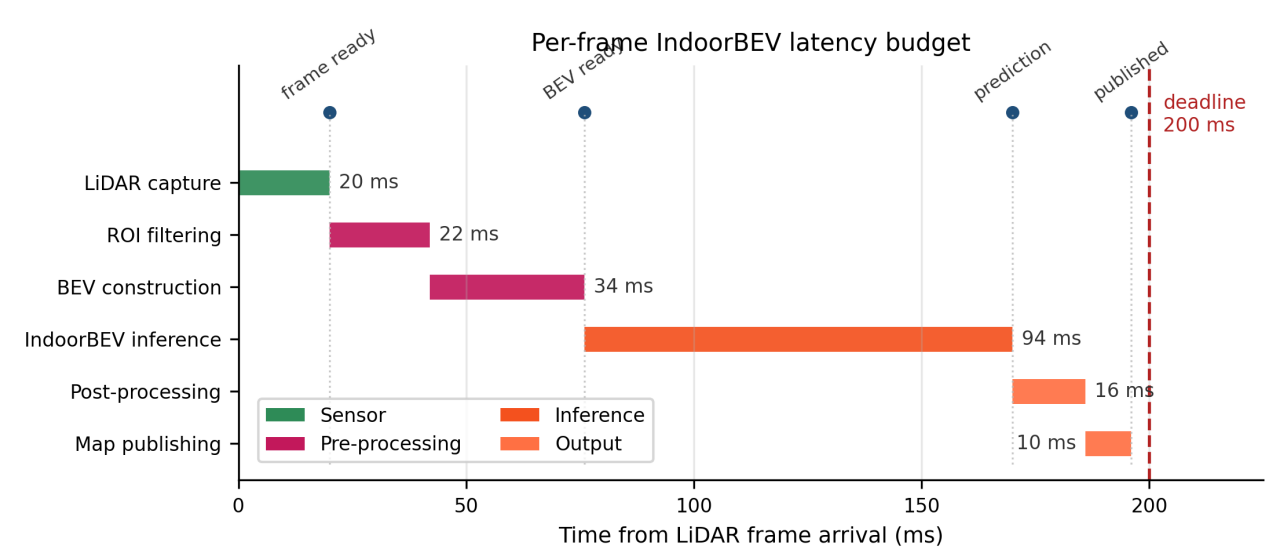

Figure 4: Per-frame latency budget of the IndoorBEV perception pipeline. Horizontal bars show the measured duration of each stage. Dotted vertical markers indicate frame readiness, BEV construction, prediction, and output publication. The dashed red line marks perception deadline.

## Height-Aware BEV Representation

Points are first filtered using a region of interest

$$\mathcal{R} = \left\{ (x,y,z) \;\middle|\; \begin{array}{l} x_{\min} \le x < x_{\max}, \quad y_{\min} \le y < y_{\max}, \\ z_{\min} \le z < z_{\max} \end{array} \right\}.$$

Each retained point is assigned to a BEV cell according to

$$u_i = \left\lfloor \frac{x_i - x_{\min}}{\Delta x} \right\rfloor, \qquad v_i = \left\lfloor \frac{y_i - y_{\min}}{\Delta y} \right\rfloor,$$

where $\Delta x$ and $\Delta y$ denote the grid resolutions.

For each occupied cell $(u,v)$, we compute

$$G_{u,v} = \left[ \mu^z_{u,v}, z^{\min}_{u,v}, z^{\max}_{u,v}, \sigma^z_{u,v}, \mu^r_{u,v}, \rho_{u,v} \right],$$

where the channels represent the mean, minimum, maximum, and standard deviation of height, mean LiDAR intensity, and normalized point density, respectively. For datasets without intensity measurements, the intensity channel is set to zero. Empty cells are assigned zero-valued features,

$$G \in \mathbb{R}^{6\times H\times W}.$$

Unlike occupancy-only BEV projection, this representation summarizes the local vertical point distribution while retaining the efficiency of 2D processing. To expose fine-grained height variations, we encode the normalized mean height $\tilde{\mu}^z_{u,v}$ using

$$\gamma(z) = [\sin(2^0\pi z), \cos(2^0\pi z), \ldots, \sin(2^{L-1}\pi z), \cos(2^{L-1}\pi z)].$$

The augmented feature map is

$$F_{\text{aug}} = \text{Concat}\left(G, \gamma(\tilde{\mu}^z)\right) \in \mathbb{R}^{(6+2L)\times H\times W},$$

where $L = 6$ in our implementation. This encoding provides additional vertical cues without constructing a dense 3D feature volume.

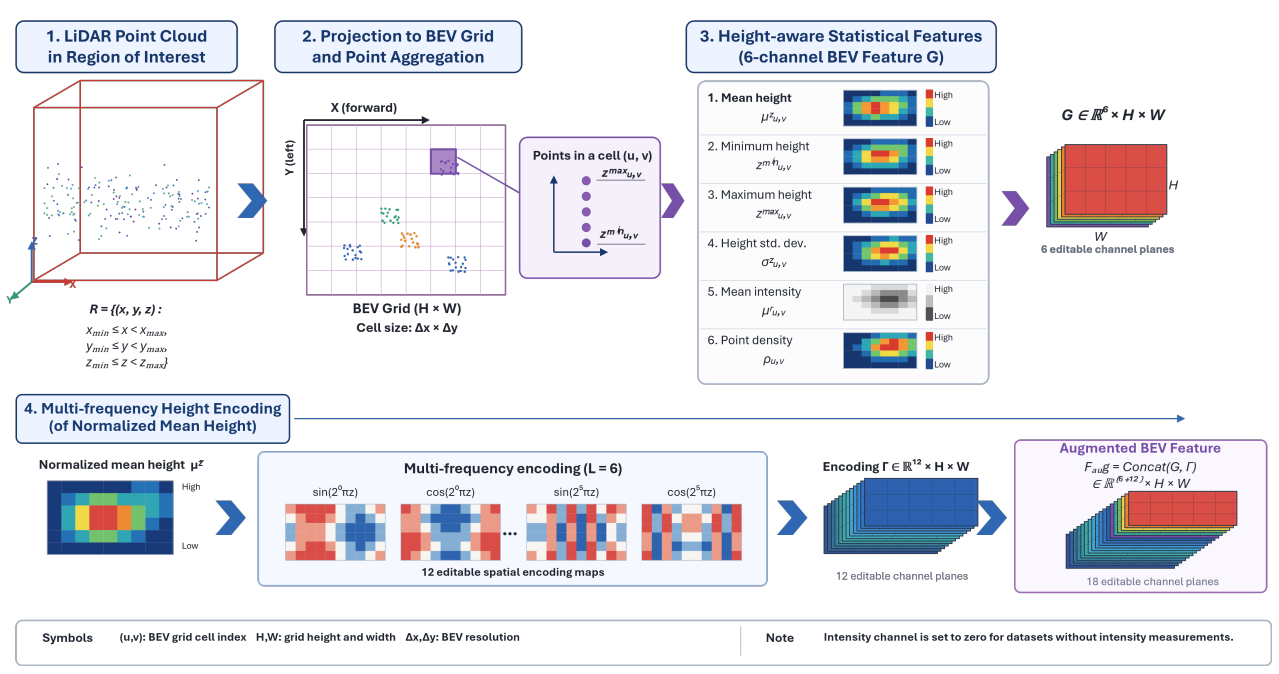

Figure 5: Height-aware BEV construction. LiDAR points within the region of interest are aggregated into an $H \times W$ grid using four height statistics, mean intensity, and normalized point density, producing $G \in \mathbb{R}^{6\times H\times W}$. Multi-frequency encoding ($L = 6$) augments the normalized mean height with 12 channels, yielding $F_{\text{aug}} \in \mathbb{R}^{(6+2L)\times H\times W}$.

## Axis-Conditioned Feature Fusion

The augmented BEV representation is processed using three parallel axis-conditioned convolutional branches:

$$F_k = \phi_k(F_{\text{aug}} \odot m_k), \qquad k \in \{1, 2, 3\},$$

where $\phi_k$ denotes a lightweight convolutional transformation and $m_k$ is a coordinate-dependent modulation map. All branches operate on the same height-aware BEV representation, but use independent parameters and coordinate conditioning to learn complementary feature transformations.

The branch outputs are adaptively aggregated as

$$F_{\text{fuse}} = \sum_{k=1}^{3} w_k F_k, \qquad \sum_{k=1}^{3} w_k = 1,$$

where the fusion weights are derived from branch-level feature statistics and geometric consistency. The fused representation is subsequently refined by parallel convolutions with dilation rates 1, 2, and 4:

$$F_{\text{enc}} = \psi_{\text{ms}}(F_{\text{fuse}}) \in \mathbb{R}^{C_e\times H\times W}.$$

This multi-scale refinement enlarges the receptive field without requiring a deep 3D backbone.

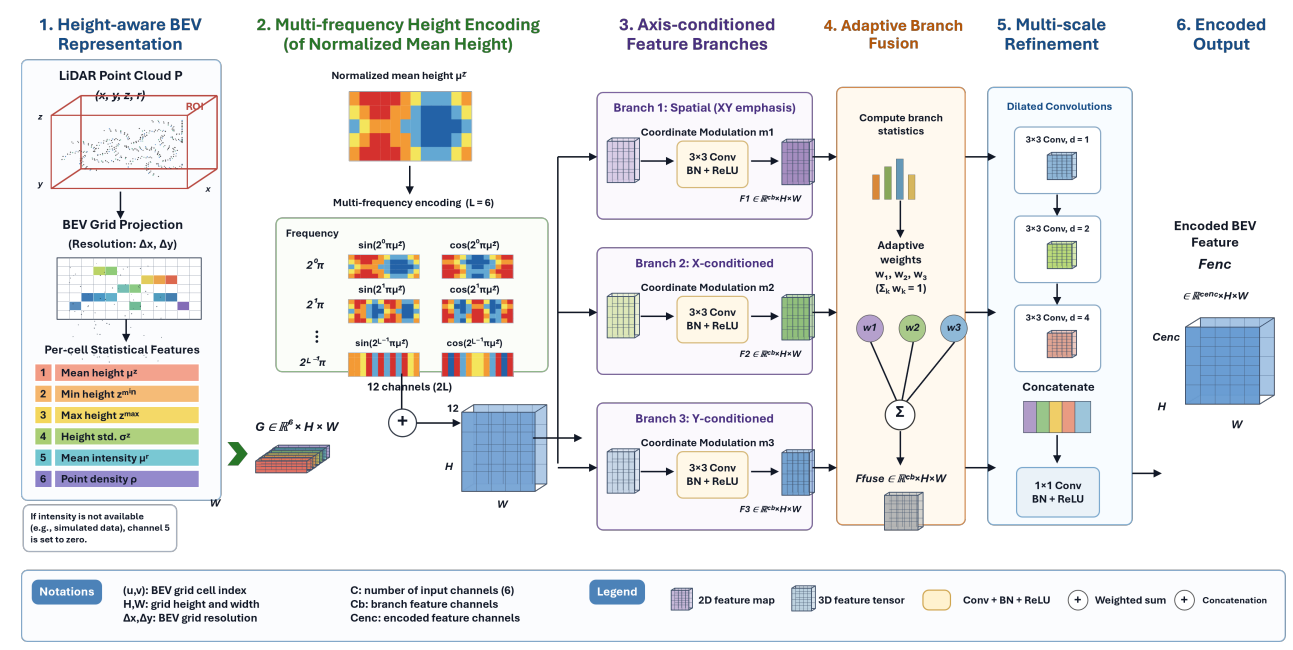

Figure 6: Axis-conditioned feature encoder. The six-channel height-aware BEV representation is augmented with a multi-frequency encoding ($L = 6$) of normalized mean height. Three coordinate-modulated branches learn complementary transformations of the same augmented feature map, which are adaptively fused and refined using dilated convolutions with rates 1, 2, and 4. A final $1 \times 1$ convolution produces $F_{\text{enc}} \in \mathbb{R}^{C_{\text{enc}}\times H\times W}$. Unavailable intensity measurements are represented by a zero-valued channel.

## Global-Local Feature Fusion

The backbone combines multi-scale spatial features with compact global scene context. A lightweight feature pyramid extracts local representations, while spatial average pooling and learned embeddings provide global context:

$$F_{\text{local}} = \text{FPN}(F_{\text{enc}}),\ p = \text{AvgPool}(F_{\text{enc}}),\ T = E + \text{FC}(p).$$

where $E \in \mathbb{R}^{N_t\times C_t}$ denotes the learned token embeddings. Residual convolutional blocks and squeeze-and-excitation modules combine multiple spatial scales and recalibrate channel responses. The aggregated token representation is projected and broadcast to the BEV resolution as $T_{\text{map}}$. The final feature map is

$$F_{\text{out}} = \text{BN}\left(\text{SE}\left(\text{Conv}_{1\times1}\left(\text{Concat}(F_{\text{local}}, T_{\text{map}})\right)\right)\right).$$

This design introduces global scene context without applying spatial self-attention over the full BEV grid.

## Prediction Targets and Training Objectives

IndoorBEV uses decoupled classification and regression heads. The classification head predicts per-cell object-class logits and probabilities:

$$Z = h_{\text{cls}}(F_{\text{out}}) \in \mathbb{R}^{C\times H\times W}, \quad S_{c,u,v} = \text{Softmax}(Z)_{c,u,v}.$$

The regression head predicts oriented bounding-box:

$$B = h_{\text{reg}}(F_{\text{out}}) \in \mathbb{R}^{5\times H\times W},\ B_{:,u,v} = (\Delta x, \Delta y, \log l, \log w, \theta).$$

Here, $(\Delta x, \Delta y)$ are offsets from a BEV cell to the object center, $(l, w)$ are the object dimensions in the BEV grid, and $\theta$ is the object orientation.

For an object of class $c$ centered at $(x_c, y_c)$, we define a Gaussian-shaped support function

$$Q_{x,y} = \exp\left(-\frac{(x-x_c)^2 + (y-y_c)^2}{2\sigma^2}\right),$$

where

$$\sigma = \max\left(\frac{\max(l_g, w_g)}{6}, 1\right),$$

and $l_g$ and $w_g$ denote the object dimensions in grid units. The categorical target is defined as

$$Y_{x,y} = \begin{cases} c, & Q_{x,y} > \eta, \\ 0, & \text{otherwise}, \end{cases}$$

where class 0 represents the background and $\eta = 0.1$. When multiple object support regions overlap, the label-generation procedure retains the assignment produced by the adopted annotation order. Bounding-box regression is supervised over the corresponding object support region:

$$\Omega_{\text{reg}} = \{(x, y) \mid Q_{x,y} > \eta\}.$$

For every $(x, y) \in \Omega_{\text{reg}}$, the offsets are measured from that cell to the continuous object center, while object dimensions are represented in logarithmic grid units.

The classification objective applies focal modulation to categorical cross-entropy:

$$\mathcal{L}_{\text{cls}} = \frac{1}{HW} \sum_{x,y} \alpha \left(1 - p_{t,x,y}\right)^{\gamma} \text{CE}(Z_{x,y}, Y_{x,y}),$$

where $p_{t,x,y}$ is the predicted probability of the target class. We use $\alpha = 2$ and $\gamma = 4$.

The regression objective is a masked Smooth L1 loss:

$$\mathcal{L}_{\text{reg}} = \frac{\sum_{(x,y)\in\Omega_{\text{reg}}} \sum_{d=1}^{5} \text{SmoothL1}\left(\hat{B}_{d,x,y} - B_{d,x,y}\right)}{|\Omega_{\text{reg}}| + \epsilon}.$$

The complete training objective is

$$\mathcal{L} = \mathcal{L}_{\text{cls}} + \lambda_{\text{reg}} \mathcal{L}_{\text{reg}},$$

where $\lambda_{\text{reg}} = 1$ for the reported IndoorBEV configuration.

## Embedded Perception Pipeline

IndoorBEV is integrated into a ROS2 perception pipeline that separates sensor reception, GPU inference, output publication, and diagnostics. LiDAR frames are stored in a bounded latest-frame queue, allowing the perception callback to process the newest available observation without accumulating stale frames. The critical path consists of frame selection, ROI filtering, BEV construction, neural-network inference, output decoding, and publication. Visualization, logging, and diagnostics are executed outside it. Details are in Sec. .

# Experiments

We evaluate IndoorBEV from two complementary perspectives: perception quality and deployment efficiency. Perception quality covers dense object-class prediction, object detection, and oriented bounding-box localization, while deployment efficiency considers model size, inference memory, latency distributions, and deadline compliance.

## Experimental Setup

**Implementation Details.** IndoorBEV is implemented in PyTorch and trained for up to 500 epochs using Adam. Unless otherwise stated, we use a batch size of 2, an initial learning rate of $10^{-4}$, and a weight decay of $10^{-4}$. The learning rate is reduced by a factor of 0.1 every 30 epochs. The classification objective uses focal loss with $\alpha = 2$ and $\gamma = 4$, while the oriented-box parameters are optimized using masked Smooth L1 loss. The classification and regression losses are equally weighted. The input point clouds are converted into six-channel BEV maps containing the mean, minimum, maximum, and standard deviation of height, mean intensity, and normalized point density. For datasets without intensity measurements, including the simulated dataset, the intensity channel is set to zero while retaining the same six-channel input interface. All experiments use a $400 \times 400$ BEV grid. The simulated and real-robot datasets use $[x_{\min}, x_{\max}] = [y_{\min}, y_{\max}] = [-2, 2]$ m and $[z_{\min}, z_{\max}] = [-1, 2]$ m, corresponding to a horizontal grid resolution of $0.01$ m per cell. Dataset-specific settings are provided in the supplementary material.

**Compared Methods.** We compare IndoorBEV with four representative point-cloud perception architectures: PointNet++, PillarNet, BEV-MAE, and GroupFree3D. PointNet++ uses hierarchical point-based feature aggregation, PillarNet performs pillar-based BEV detection, BEV-MAE learns transformer-based BEV representations, and GroupFree3D is a proposal-free indoor 3D detector. The baseline architectures are adapted to the common dense BEV output space using task-specific classification and regression heads. Within each dataset, all methods use the same training, validation, and test splits, spatial ranges, class definitions, and evaluation protocols. The checkpoint with the lowest validation loss is used for evaluation.

For compactness, we denote PillarNet, PointNet++, BEV-MAE, GroupFree3D, and IndoorBEV as M1, M2, M3, M4, and M5, respectively, in the detailed result tables.

**Embedded Platform.** Runtime experiments are conducted on an NVIDIA AGX Orin with batch size one. All methods are evaluated using the same numerical precision, software environment, input samples, and timing boundaries. GPU execution is synchronized at each timing endpoint. The detailed warm-up procedure, number of evaluated frames, and memory-measurement protocol are in the Appendix.

**Latency Protocol.** For frame $i$, let $t_i^{\text{cap}}$ denote the sensor capture time, $t_i^{\text{start}}$ the start of perception processing, and $t_i^{\text{pub}}$ the output publication time. We define end-to-end latency and processing latency as

$$T_i^{\text{e2e}} = t_i^{\text{pub}} - t_i^{\text{cap}}, \qquad T_i^{\text{proc}} = t_i^{\text{pub}} - t_i^{\text{start}}. \quad (1)$$

The former measures the age of the sensor observation delivered to downstream modules, whereas the latter measures execution time after the frame is selected for processing.

Deadline compliance is evaluated using $T_i^{\text{e2e}}$. Given a target deadline $D = 200$ ms, the deadline miss ratio is

$$\text{MissRatio} = \frac{1}{N} \sum_{i=1}^{N} \mathbb{I}\left[T_i^{\text{e2e}} > D\right]. \quad (2)$$

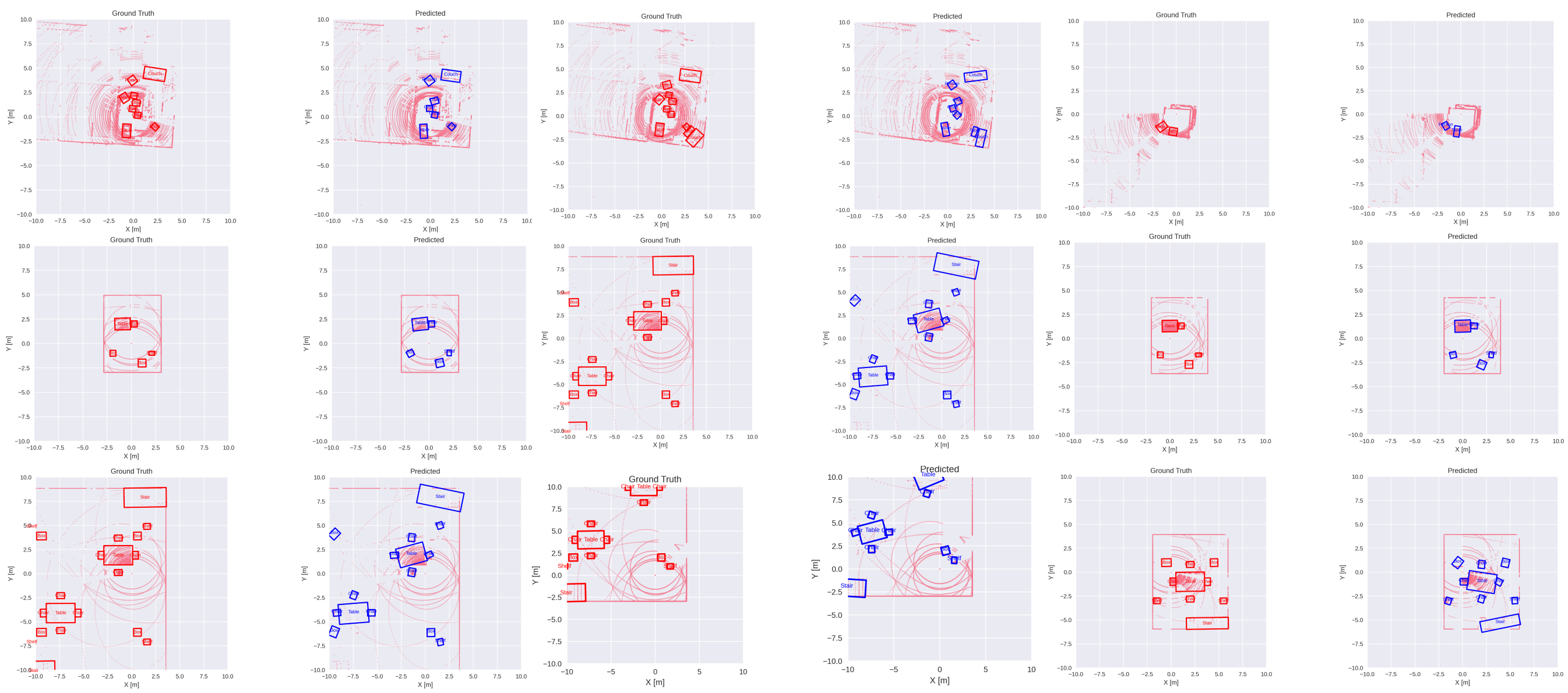

Figure 7: Qualitative comparison of IndoorBEV detection results across real-world and simulated environments. Each row shows a different scene with ground truth (left) and predicted (right) oriented bounding boxes (red) and BEV heatmaps. First column presents results from real-world LiDAR scans, while subsequent columns demonstrate performance on synthetic data with varying indoor layouts and object configurations. Blue boxes indicate model predictions and red boxes show ground truth.

We report the mean, median, 95th percentile, 99th percentile, and maximum latency. The 200 ms deadline corresponds to the target period of a 5 Hz perception pipeline.

Table 1: Aggregate perception performance across the three evaluation datasets. Best results are shown in bold.

| Dataset | Method | Acc. | mAP | mIoU | Acc@.50 |
|---|---|---|---|---|---|
| Simulation | PillarNet | 0.62 | 0.47 | 0.66 | 0.74 |
| | PointNet++ | 0.64 | 0.49 | 0.70 | 0.81 |
| | BEV-MAE | 0.50 | 0.42 | 0.79 | 0.87 |
| | GroupFree3D | 0.65 | 0.53 | 0.68 | 0.79 |
| | IndoorBEV | **0.86** | **0.90** | **0.86** | **0.97** |
| Real World | PillarNet | 0.44 | 0.33 | **0.81** | 0.89 |
| | PointNet++ | 0.42 | 0.34 | 0.80 | 0.90 |
| | BEV-MAE | 0.39 | 0.30 | 0.69 | 0.79 |
| | GroupFree3D | 0.41 | 0.34 | 0.76 | 0.85 |
| | IndoorBEV | **0.61** | **0.63** | 0.72 | **0.91** |
| LiDAR-Net | PillarNet | 0.25 | 0.23 | 0.55 | 0.50 |
| | PointNet++ | 0.22 | 0.21 | 0.58 | 0.55 |
| | BEV-MAE | 0.20 | 0.19 | 0.52 | 0.45 |
| | GroupFree3D | 0.23 | 0.24 | 0.57 | 0.52 |
| | IndoorBEV | **0.30** | **0.29** | **0.62** | **0.60** |

Acc@.50 denotes Acc@IoU$_{0.50}$. Detailed per-class results and regression-loss statistics are provided in the Appendix.

## Datasets and Evaluation Protocol

### Datasets

We evaluate IndoorBEV on simulated LiDAR scenes, real-world robot scans, and the public LiDAR-Net dataset (Guo et al. 2024). The simulated dataset contains table, chair, shelf, box, and stair, while the real-world dataset contains couch, table, person, and chair. Real-world annotations are generated semi-automatically and manually verified using SUSTech-POINTS (Li et al. 2020). All datasets are converted to a common BEV input and target format; detailed splits, class mappings, spatial ranges, and intensity settings are provided in the Appendix.

**Evaluation Metrics.** For dense object-class prediction, we report per-class precision, recall, and F1-score, together with overall BEV classification accuracy. For object detection, we report class-wise average precision and mean average precision. Localization quality is evaluated using mean oriented IoU and the proportions of matched predictions exceeding IoU thresholds of 0.25, 0.50, and 0.75. Detailed per-class results and regression-loss statistics are in the Appendix.

## Main Perception Results

Table 1 summarizes the classification, detection, and localization performance across the simulated, real-world, and LiDAR-Net datasets. The table reports aggregate metrics, while detailed per-class precision, recall, F1, AP, and localization results are provided in the supplementary material.

**Simulated Indoor Scenes.** IndoorBEV achieves an overall classification accuracy of 0.86 and a detection mAP of 0.90 on the simulated dataset. It also obtains a mean IoU of 0.86 and an Acc@IoU$_{0.50}$ of 0.97. The improvements are consistent across the five evaluated object categories, indicating that the height-aware representation is effective for objects with different vertical profiles and ground-plane footprints.

**Real-World Robot Scans.** On the real-world dataset, IndoorBEV achieves the highest classification accuracy and detection mAP, reaching 0.61 and 0.63, respectively. It also achieves the highest Acc@IoU$_{0.50}$ of 0.91. However, its mean IoU of 0.72 is lower than those of PillarNet and PointNet++, and its Acc@IoU$_{0.75}$ is lower under the stricter localization criterion. These results indicate that IndoorBEV improves object recognition and moderate-threshold localization, while precise boundary alignment remains more challenging in real indoor scenes.

**LiDAR-Net.** IndoorBEV achieves an overall accuracy of 0.30 and a detection mAP of 0.29 on LiDAR-Net. It also obtains the highest mean IoU of 0.62 and Acc@IoU$_{0.50}$ of 0.60 among the evaluated methods. Although the absolute performance is lower than on the simulated and robot-collected datasets, the results show that the proposed representation can be applied to an independently collected LiDAR dataset.

## Embedded Deployment Results

We evaluate model complexity, latency distributions, deadline compliance, and peak inference memory on the NVIDIA AGX Orin. The combined complexity–latency–memory comparison is shown in Fig. 1, while Table 2 reports the corresponding numerical results. IndoorBEV contains 0.54M parameters and requires 2.3 MB of model storage and 21.52 MB of peak inference memory. Its mean latency is 169.6 ms, with a p95 latency of 188.4 ms and a p99 latency of 196.7 ms. The maximum observed latency is 204.5 ms, resulting in a deadline miss ratio of 1.8%. IndoorBEV is therefore the only evaluated method whose mean and p99 latencies remain below the 200 ms target deadline. BEV-MAE has the largest resource footprint, using 121.5M parameters and 483.75 MB of peak inference memory. Relative to BEV-MAE, IndoorBEV uses 225 times fewer parameters and approximately 22.5 times less inference memory. Fig. 3 further reports the cumulative latency distribution. The distribution of IndoorBEV is concentrated below the target deadline, although a small number of frames exceed 200 ms. The baseline distributions remain entirely above the target deadline under the same runtime protocol.

Table 2: Real-time and resource comparison on the embedded AGX Orin platform. The deadline is $D = 200$ ms and all measurements use batch size one.

**(a) Model complexity and inference memory**

| Method | Params (M) | Size (MB) | GPU Mem. (MB) |
|---|---|---|---|
| IndoorBEV | **0.54** | **2.3** | **21.52** |
| PillarNet | 62.3 | 192 | 157.63 |
| PointNet++ | 42.1 | 158 | 176.40 |
| GroupFree3D | 87.4 | 228 | 325.92 |
| BEV-MAE | 121.5 | 464 | 483.75 |

**(b) End-to-end latency and deadline compliance**

| Method | Mean | Med. | p95 | p99 | Max | Miss |
|---|---|---|---|---|---|---|
| IndoorBEV | **169.6** | **166.8** | **188.4** | **196.7** | **204.5** | **1.8** |
| PillarNet | 286.0 | 281.5 | 326.8 | 348.2 | 371.6 | 100.0 |
| PointNet++ | 297.0 | 292.4 | 341.7 | 365.3 | 389.8 | 100.0 |
| GroupFree3D | 372.0 | 366.9 | 423.5 | 451.2 | 487.6 | 100.0 |
| BEV-MAE | 454.0 | 448.6 | 512.4 | 548.9 | 586.7 | 100.0 |

## Ablation Study

We analyze the contribution of the major IndoorBEV components and the accuracy–resource tradeoff of efficiency-oriented model variants on the simulated indoor dataset. The current ablation results are summarized in Table 3. Removing global scene tokens reduces the mAP from 0.90 to 0.87 and the mean IoU from 0.86 to 0.83, indicating that compact scene-level context contributes to both detection and localization. Removing the axis-conditioned fusion module causes larger reductions, producing an mAP of 0.84 and a mean IoU of 0.80. This result supports the use of multiple coordinate-conditioned feature transformations for processing the height-aware BEV representation. The lightweight-backbone and smaller-grid configurations further reduce latency and memory consumption, but at the cost of lower perception accuracy. In particular, the smaller-grid variant reduces latency to 109.8 ms and memory usage to 12.84 MB, while its mAP decreases to 0.78. These variants provide alternative operating points when the robot has a stricter computational budget. Additional component ablations for height encoding and multi-scale refinement will be reported in the supplementary material.

Table 3: Accuracy–resource tradeoff under different IndoorBEV configurations on the simulated indoor dataset.

| Configuration | mAP | mIoU | Latency | Memory |
|---|---|---|---|---|
| Full IndoorBEV | **0.90** | **0.86** | 169.6 | 21.52 |
| w/o global tokens | 0.87 | 0.83 | 163.4 | 20.18 |
| w/o axis-cond. fusion | 0.84 | 0.80 | 158.7 | 19.64 |
| Lightweight backbone | 0.82 | 0.78 | 132.5 | 16.37 |
| Smaller BEV grid | 0.78 | 0.73 | **109.8** | **12.84** |

## Qualitative Results

Fig. 7 presents representative predictions from the simulated and real-world datasets. IndoorBEV generally captures object locations and scene layouts while maintaining spatial consistency between the dense object-class map and oriented bounding boxes. Errors are more frequent in heavily occluded regions, for objects with ambiguous boundaries. These observations are consistent with the lower Acc@IoU$_{0.75}$ obtained on the real-world dataset.

# Conclusion

We presented IndoorBEV, a lightweight LiDAR-based BEV perception system for real-time indoor mobile robots. By combining height-aware BEV feature construction, multi-axis geometric fusion, compact global-local context modeling, and decoupled dense prediction heads, IndoorBEV provides semantic scene understanding and oriented object localization with low memory and bounded inference latency.

## References

Armeni, I.; Sener, O.; Zamir, A. R.; Jiang, H.; Brilakis, I.; Fischer, M.; and Savarese, S. 2016. 3D Semantic Parsing of Large-Scale Indoor Spaces. In *Proceedings of the IEEE Conference on Computer Vision and Pattern Recognition (CVPR)*, 1534–1543.

Bédard, C.; Dagenais, M.; et al. 2022. ros2_tracing: Tracing for ROS 2. In *Proceedings of the ROSCon Conference*.

Cao, A.-Q.; and De Charette, R. 2022. MonoScene: Monocular 3D Semantic Scene Completion. In *Proceedings of the IEEE/CVF Conference on Computer Vision and Pattern Recognition (CVPR)*, 3991–4001.

Casini, D.; Blaß, T.; Lütkebohle, I.; and Brandenburg, B. 2019. Response-Time Analysis of ROS 2 Processing Chains Under Reservation-Based Scheduling.

Chang, A. X.; Dai, A.; Funkhouser, T.; Halber, M.; Niessner, M.; Savva, M.; Song, S.; Zeng, A.; and Zhang, Y. 2017. Matterport3D: Learning from RGB-D Data in Indoor Environments. In *Proceedings of the International Conference on 3D Vision (3DV)*, 667–676.

Chen, S.; et al. 2023. BEV-MAE: Bird's-Eye-View Masked Autoencoders for Self-supervised Pre-training of 3D Object Detectors. In *Proceedings of the IEEE/CVF International Conference on Computer Vision (ICCV)*.

Chen, T.; Moreau, T.; Jiang, Z.; Zheng, L.; Yan, E.; Shen, H.; Cowan, B.; Wang, L.; Hu, Y.; Ceze, L.; et al. 2018. TVM: An Automated End-to-End Optimizing Compiler for Deep Learning. In *Proceedings of the 13th USENIX Symposium on Operating Systems Design and Implementation (OSDI)*, 578–594.

Chen, X.; Ma, H.; Wan, J.; Li, B.; and Xia, T. 2017. Multi-View 3D Object Detection Network for Autonomous Driving. In *Proceedings of the IEEE Conference on Computer Vision and Pattern Recognition (CVPR)*, 1907–1915.

Choy, C.; Gwak, J.; and Savarese, S. 2019. 4D Spatio-Temporal ConvNets: Minkowski Convolutional Neural Networks. In *Proceedings of the IEEE/CVF Conference on Computer Vision and Pattern Recognition (CVPR)*, 3075–3084.

Dai, A.; Chang, A. X.; Savva, M.; Halber, M.; Funkhouser, T.; and Nießner, M. 2017. ScanNet: Richly-Annotated 3D Reconstructions of Indoor Scenes. In *Proceedings of the IEEE Conference on Computer Vision and Pattern Recognition (CVPR)*, 5828–5839.

Graham, B.; Engelcke, M.; and van der Maaten, L. 2018. Submanifold Sparse Convolutional Networks. In *Proceedings of the IEEE/CVF Conference on Computer Vision and Pattern Recognition (CVPR)*, 9224–9232.

Guo, Y.; Li, Y.; Ren, D.; Zhang, X.; Li, J.; Pu, L.; Ma, C.; Zhan, X.; Guo, J.; Wei, M.; et al. 2024. LiDAR-Net: A Real-Scanned 3D Point Cloud Dataset for Indoor Scenes. In *Proceedings of the IEEE/CVF Conference on Computer Vision and Pattern Recognition (CVPR)*.

Han, S.; Pool, J.; Tran, J.; and Dally, W. J. 2015. Learning Both Weights and Connections for Efficient Neural Networks. *Advances in Neural Information Processing Systems (NeurIPS)*, 28.

Hinton, G.; Vinyals, O.; and Dean, J. 2015. Distilling the Knowledge in a Neural Network. *arXiv preprint arXiv:1503.02531*.

Hu, Q.; Yang, B.; Xie, L.; Rosa, S.; Guo, Y.; Wang, Z.; Trigoni, N.; and Markham, A. 2020. RandLA-Net: Efficient Semantic Segmentation of Large-Scale Point Clouds. In *Proceedings of the IEEE/CVF Conference on Computer Vision and Pattern Recognition (CVPR)*, 11108–11117.

Jacob, B.; Kligys, S.; Chen, B.; Zhu, M.; Tang, M.; Howard, A.; Adam, H.; and Kalenichenko, D. 2018. Quantization and Training of Neural Networks for Efficient Integer-Arithmetic-Only Inference. In *Proceedings of the IEEE/CVF Conference on Computer Vision and Pattern Recognition (CVPR)*, 2704–2713.

Lang, A. H.; Vora, S.; Caesar, H.; Zhou, L.; Yang, J.; and Beijbom, O. 2019. PointPillars: Fast Encoders for Object Detection from Point Clouds. In *Proceedings of the IEEE/CVF Conference on Computer Vision and Pattern Recognition (CVPR)*, 12697–12705.

Li, E.; Wang, S.; Li, C.; Li, D.; Wu, X.; and Hao, Q. 2020. SUSTech POINTS: A Portable 3D Point Cloud Interactive Annotation Platform System. In *2020 IEEE Intelligent Vehicles Symposium (IV)*, 1108–1115.

Li, Y.; Yu, Y.; Liao, Y.; Zhang, S.; Lu, W.; and Yang, J. 2022a. BEVDepth: Acquisition of Reliable Depth for Multi-View 3D Object Detection. In *Proceedings of the AAAI Conference on Artificial Intelligence*, volume 36, 1483–1491.

Li, Z.; Wang, W.; Li, H.; Xie, E.; Sun, M.; Li, S.; Yu, Z.; Luo, S.; Zhang, T.; Li, Y.; et al. 2022b. BEVFormer: Learning Bird's-Eye-View Representation from Multi-Camera Images via Spatiotemporal Transformers. In *Proceedings of the European Conference on Computer Vision (ECCV)*, 1–18.

Liu, Z.; Zhang, Z.; Cao, Y.; Hu, H.; and Tong, X. 2021. Group-Free 3D Object Detection via Transformers. In *2021 IEEE/CVF International Conference on Computer Vision (ICCV)*, 2949–2958. IEEE.

Liu, Z.; Zhou, Y.; Zhou, Y.; Bai, X.; et al. 2023. BEVFusion: Multi-Task Multi-Sensor Fusion with Unified Bird's-Eye View Representation. In *Proceedings of the IEEE/CVF Conference on Computer Vision and Pattern Recognition (CVPR)*, 2705–2715.

Macenski, S.; Foote, T.; Gerkey, B.; Lalancette, C.; and Woodall, W. 2022. Robot Operating System 2: Design, Architecture, and Uses in the Wild. *Science Robotics*, 7(66).

Maruyama, Y.; Kato, S.; and Azumi, T. 2016. Exploring the Performance of ROS2. In *Proceedings of the International Conference on Embedded Software (EMSOFT)*, 1–10.

Micikevicius, P.; Narang, S.; Alben, J.; Diamos, G.; Elsen, E.; Garcia, D.; Ginsburg, B.; Houston, M.; Kuchaiev, O.; Venkatesh, G.; and Wu, H. 2017. Mixed Precision Training. *arXiv preprint arXiv:1710.03740*.

Misra, I.; et al. 2021. An End-to-End Transformer Model for 3D Object Detection. In *Proceedings of the IEEE/CVF International Conference on Computer Vision (ICCV)*.

NVIDIA Corporation. ???? NVIDIA TensorRT. https://developer.nvidia.com/tensorrt. Accessed: 2026-05-23.

ONNX Runtime Contributors. ???? ONNX Runtime: High-Performance Inference Engine. https://onnxruntime.ai. Accessed: 2026-05-23.

Qi, C. R.; Chen, X.; Litany, O.; and Guibas, L. J. 2020. ImVoteNet: Boosting 3D Object Detection in Point Clouds with Image Votes. In *Proceedings of the IEEE/CVF Conference on Computer Vision and Pattern Recognition (CVPR)*, 4404–4413.

Qi, C. R.; Litany, O.; He, K.; and Guibas, L. J. 2019. VoteNet: A Deep Learning Labeling Approach for 3D Object Detection in Point Clouds. In *Proceedings of the IEEE/CVF International Conference on Computer Vision (ICCV)*, 6201–6210.

Qi, C. R.; Su, H.; Mo, K.; and Guibas, L. J. 2017a. PointNet: Deep Learning on Point Sets for 3D Classification and Segmentation. *Proceedings of the IEEE Conference on Computer Vision and Pattern Recognition (CVPR)*, 652–660.

Qi, C. R.; Yi, L.; Su, H.; and Guibas, L. J. 2017b. PointNet++: Deep Hierarchical Feature Learning on Point Sets in a Metric Space. In *Advances in Neural Information Processing Systems (NeurIPS)*, volume 30.

Quigley, M.; Conley, K.; Gerkey, B.; Faust, J.; Foote, T.; Leibs, J.; Wheeler, R.; and Ng, A. Y. 2009. ROS: An Open-Source Robot Operating System. In *Proceedings of the IEEE International Conference on Robotics and Automation (ICRA) Workshop on Open Source Software*.

Rukhovich, D.; Vorontsova, A.; and Konushin, A. 2022. FCAF3D: Fully Convolutional Anchor-Free 3D Object Detection. In *Proceedings of the European Conference on Computer Vision (ECCV)*.

Shi, S.; Guo, C.; Jiang, L.; Wang, Z.; Shi, J.; Wang, X.; and Li, H. 2022. PillarNet: Real-time and High-performance Pillar-based 3D Object Detection. In *Proceedings of the European Conference on Computer Vision (ECCV)*, 35–52.

Silberman, N.; Hoiem, D.; Kohli, P.; and Fergus, R. 2012. Indoor Segmentation and Support Inference from RGBD Images. *Proceedings of the European Conference on Computer Vision (ECCV)*, 746–760.

Song, S.; Lichtenberg, S. P.; and Xiao, J. 2015. SUN RGB-D: A RGB-D Scene Understanding Benchmark Suite. In *Proceedings of the IEEE Conference on Computer Vision and Pattern Recognition (CVPR)*, 567–576.

Teng, Z.; Zhang, J.; Yang, K.; Peng, K.; Shi, H.; Reiß, S.; Cao, K.; and Stiefelhagen, R. 2024. 360BEV: Panoramic Semantic Mapping for Indoor Bird's-Eye View. In *Proceedings of the IEEE/CVF Winter Conference on Applications of Computer Vision (WACV)*.

Thomas, H.; Qi, C. R.; Deschaud, J.-E.; Marcotegui, B.; Goulette, F.; and Guibas, L. J. 2019. KPConv: Flexible and Deformable Convolution for Point Clouds. In *Proceedings of the IEEE/CVF International Conference on Computer Vision (ICCV)*, 6411–6420.

Wang, J.; et al. 2022. CAGroup3D: Class-Aware Grouping for 3D Object Detection. In *Proceedings of the European Conference on Computer Vision (ECCV)*.

Wang, Y.; Sun, Y.; Liu, Z.; Sarma, S. E.; Bronstein, M. M.; and Solomon, J. M. 2019. Dynamic Graph CNN for Learning on Point Clouds. *ACM Transactions on Graphics (TOG)*, 38(5): 1–12.

Xu, B.; Tao, A.; Thomas, H.; Zhang, J.; and Barfoot, T. D. 2025a. MakeWay: Object-Aware Costmaps for Proactive Indoor Navigation Using LiDAR. In *Proceedings of the Conference on Robots and Vision (CRV)*. Preprint available 2024.

Xu, Z.; Shen, H.; Han, X.; Jin, H.; Ye, K.; and Shimada, K. 2025b. LV-DOT: LiDAR-Visual Dynamic Obstacle Detection and Tracking for Autonomous Robot Navigation. *arXiv preprint arXiv:2502.20607*.

Yan, Y.; Mao, Y.; and Li, B. 2018. SECOND: Sparsely Embedded Convolutional Detection. *Sensors*, 18(10): 3337.

Yang, B.; Luo, W.; and Urtasun, R. 2018. PIXOR: Real-time 3D Object Detection from Point Clouds. In *Proceedings of the IEEE Conference on Computer Vision and Pattern Recognition (CVPR)*, 7652–7660.

Ye, W.; et al. 2023. Real-Time Performance Evaluation of ROS 2 with PREEMPT_RT. *IEEE Robotics and Automation Letters*.

Yin, T.; Zhou, X.; and Krahenbuhl, P. 2021. Center-based 3D Object Detection and Tracking. In *Proceedings of the IEEE/CVF Conference on Computer Vision and Pattern Recognition (CVPR)*, 11784–11793.

Zhang, Z.; Sun, Y.; Yang, S.; Chen, X.; Xu, L.; Wang, L.; Cai, J.; and Zhao, J. 2020. H3DNet: 3D Object Detection Using Hybrid Geometric Primitives. In *Proceedings of the European Conference on Computer Vision (ECCV)*, 311–329.

Zhao, H.; Jiang, L.; Jia, J.; Torr, P. H.; and Koltun, V. 2021. Point Transformer. In *Proceedings of the IEEE/CVF International Conference on Computer Vision (ICCV)*, 16259–16268.

Zhou, H.; Qi, L.; Wang, J.; Ma, Y.; Hao, P.; Yu, Q.; Yang, R.; and Ma, L. 2021. Panoptic-PolarNet: Proposal-free LiDAR Point Cloud Panoptic Segmentation. In *Proceedings of the IEEE/CVF Conference on Computer Vision and Pattern Recognition (CVPR)*, 13194–13203.

Zhou, Y.; and Tuzel, O. 2018. VoxelNet: End-to-End Learning for Point Cloud Based 3D Object Detection. In *Proceedings of the IEEE Conference on Computer Vision and Pattern Recognition (CVPR)*, 4490–4499.

Zhu, X.; Zhou, J.; Wang, Y.; Li, Z.; Li, J.; Liu, Y.; and Li, Y. 2021. Cylindrical and Asymmetrical 3D Convolution Networks for LiDAR Segmentation. In *Proceedings of the IEEE/CVF Conference on Computer Vision and Pattern Recognition (CVPR)*, 9939–9948.

# Appendix

## Additional Experimental Details

This appendix provides additional implementation details, dataset and target-generation protocols, baseline adaptation procedures, complete per-class results, component ablations, a further study of orientation representation, and supplementary embedded-system analysis. In the detailed comparison tables, PillarNet, PointNet++, BEV-MAE, GroupFree3D, and IndoorBEV are denoted as M1, M2, M3, M4, and M5, respectively. All primary M1–M5 comparisons and the component-ablation experiments use the initial direct-angle IndoorBEV formulation reported in the main paper. The subsequently developed double-angle formulation is evaluated separately in section."Extended Double-Angle Ablation Analysis.".

### Implementation and Training Details

IndoorBEV is implemented in PyTorch and trained using the Adam optimizer. For the simulated and real-world datasets, training is performed for up to 500 epochs with a batch size of 4, an initial learning rate of $10^{-4}$, and a weight decay of $10^{-4}$. A StepLR schedule reduces the learning rate by a factor of $0.1$ every 30 epochs. The checkpoint with the lowest validation loss is used for evaluation. Early stopping is enabled after a 30-epoch warm-up period when the validation loss does not improve by at least $10^{-4}$ for 45 consecutive epochs. All reported training runs use a fixed random seed of 42.

The classification objective applies focal modulation to categorical cross-entropy:

$$\mathcal{L}_{\text{cls}} = \frac{1}{HW} \sum_{x,y} \alpha \left(1 - p_{t,x,y}\right)^{\gamma} \text{CE}(Z_{x,y}, Y_{x,y}), \qquad (3)$$

where $Z_{x,y}$ denotes the predicted class logits, $Y_{x,y}$ denotes the categorical target, and $p_{t,x,y}$ is the predicted probability of the target class. We use $\alpha = 2$ and $\gamma = 4$.

The direct-angle IndoorBEV formulation used for the primary comparisons and ablation experiments predicts five regression channels:

$$B_{x,y} = \left(\Delta x, \Delta y, \log l, \log w, \theta\right). \qquad (4)$$

Bounding-box regression uses masked Smooth L1 loss:

$$\mathcal{L}_{\text{reg}} = \frac{\sum_{(x,y)\in\Omega_{\text{reg}}} \sum_{d=1}^{5} \text{SmoothL1}\left(\hat{B}_{d,x,y} - B_{d,x,y}\right)}{|\Omega_{\text{reg}}| + \epsilon}. \qquad (5)$$

Unless otherwise specified, the classification and regression objectives are equally weighted:

$$\mathcal{L} = \mathcal{L}_{\text{cls}} + \mathcal{L}_{\text{reg}}. \qquad (6)$$

A further orientation study in section."Extended Double-Angle Ablation Analysis." evaluates a double-angle representation motivated by the geometric semantics of principal-axis perception in robotics. That study changes only the orientation representation and its associated loss; the remaining IndoorBEV architecture is unchanged.

### BEV Input Representation and Spatial Ranges

Each point cloud is converted into a $400 \times 400$ BEV grid. The full representation contains six channels:

$$\left(\bar{z}, z_{\min}, z_{\max}, \sigma_z, \bar{r}, \rho\right), \qquad (7)$$

where $\bar{z}$, $z_{\min}$, $z_{\max}$, and $\sigma_z$ denote the mean, minimum, maximum, and standard deviation of point height within a cell, $\bar{r}$ denotes mean intensity, and $\rho$ denotes normalized point density. For datasets without intensity measurements, the intensity channel is set to zero while retaining the same six-channel interface.

The simulated dataset uses

$$[x_{\min}, x_{\max}] = [y_{\min}, y_{\max}] = [-10, 10]\,\text{m}, \quad [z_{\min}, z_{\max}] = [-1, 2]\,\text{m}. \qquad (8)$$

The resulting horizontal resolution is $0.05$ m per cell. The real-world dataset uses

$$[x_{\min}, x_{\max}] = [y_{\min}, y_{\max}] = [-2, 2]\,\text{m}, \quad [z_{\min}, z_{\max}] = [-1, 2]\,\text{m}, \qquad (9)$$

which corresponds to $0.01$ m per cell.

The available samples are divided into training, validation, and test subsets using a fixed 80/10/10 split protocol with random seed 42. The same stored split files are reused by all compared methods and ablation configurations. The simulated and real-world test splits contain 255 and 52 frames, respectively.

### Dataset Details

**Simulated indoor dataset.** The simulated benchmark contains indoor scenes with varying layouts, object arrangements, and clutter levels. The evaluated categories are table, chair, shelf, box, and stair. The simulated point clouds do not provide intensity measurements, so the intensity channel is set to zero. The test split contains 2,565 annotated object instances, comprising 329 tables, 1,068 chairs, 444 shelves, 552 boxes, and 172 stairs.

**Real-world robot dataset.** The real-world point clouds are collected using a 3D LiDAR mounted on a mobile robot. The evaluated classes are couch, table, person, and chair. The scenes include clutter, partial occlusion, moving agents, and large variation in point support. Ground-truth annotations are generated using a semi-automatic annotation procedure and manually inspected with SUSTechPOINTS. The geometric consistency of center, full object extent, and principal axis is more difficult to maintain in partially observed real scenes than in simulation. This issue is considered in the orientation study and limitations discussion.

### Target Generation Details

For each object of class $c$ centered at $(x_c, y_c)$, the Gaussian-shaped support function is

$$Q_{x,y} = \exp\left(-\frac{(x - x_c)^2 + (y - y_c)^2}{2\sigma^2}\right), \qquad (10)$$

where

$$\sigma = \max\left(\frac{\max(l_g, w_g)}{6}, 1\right), \qquad (11)$$

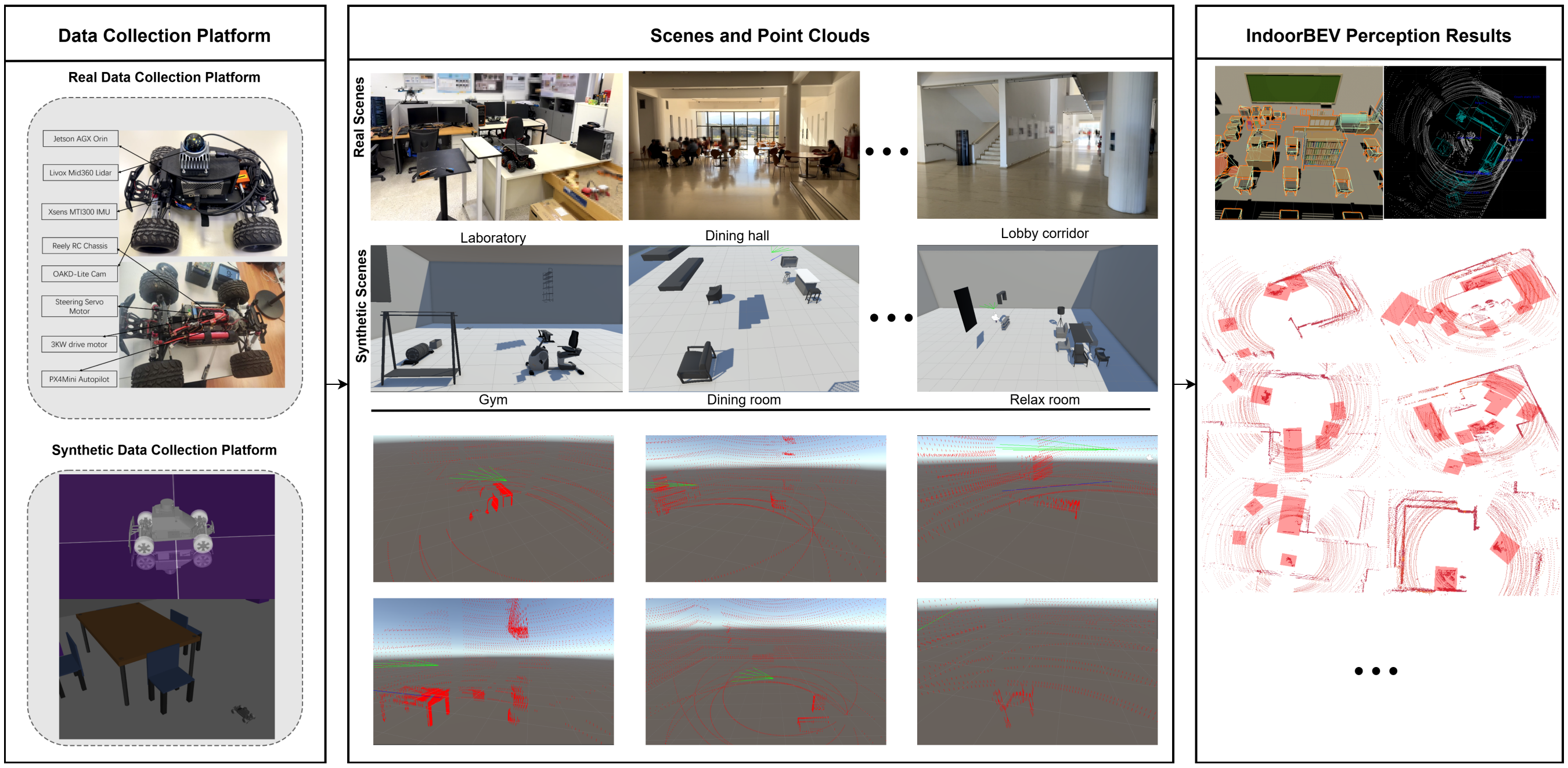

Figure 8: Processing latency under sequential and pipelined execution across input point-count intervals. Center lines denote medians, boxes show the interquartile range, and whiskers follow the convention used by the plotting implementation. The dashed horizontal line marks the $200\ \mathrm{ms}$ target deadline.

and $l_g$ and $w_g$ denote object dimensions in grid units.

The categorical BEV target is

$$Y_{x,y} = \begin{cases} c, & Q_{x,y} > \eta, \\ 0, & \text{otherwise}, \end{cases} \tag{12}$$

where class 0 denotes background and $\eta = 0.1$. Bounding-box regression is supervised over

$$\Omega_{\text{reg}} = \{(x, y) \mid Q_{x,y} > \eta\}. \tag{13}$$

For each supervised cell, the regression target contains the offset from the cell to the continuous object center, logarithmic object length and width in grid units, and object orientation. The main experiments use annotation-order assignment when object-support regions overlap.

## Baseline Adaptation Details

The evaluated baselines were originally developed for different point-cloud representations and prediction tasks. Each architecture is adapted to the common dense BEV output space and evaluation protocol used by IndoorBEV.

**PointNet++.** PointNet++ processes a fixed-size point set using hierarchical set abstraction. Each point cloud is sampled or repeated to contain 24,000 points. The resulting representation is decoded into a dense $400 \times 400$ BEV feature map using transposed convolutions, followed by task-specific classification and regression heads.

**PillarNet.** PillarNet converts point clouds into pillar-based BEV features. The adapted implementation applies classification and regression heads in the same BEV coordinate system used by IndoorBEV.

**BEV-MAE.** BEV-MAE processes the BEV input through patch embedding and transformer blocks. The resulting spatial representation is reshaped and upsampled to the common BEV resolution before the task-specific heads are applied.

**GroupFree3D.** GroupFree3D is adapted to the evaluated indoor categories and coordinate system. Its object predictions are transformed into the common BEV detection and localization format for evaluation.

Within each dataset, all methods use the same training, validation, and test splits, evaluated categories, spatial range, and metric implementations. All M1–M5 values reported in this appendix are obtained from the same test outputs used in the main paper. Model parameters, storage requirements, latency, and inference memory are measured using the actual adapted implementations rather than values copied from the original publications. Predicted boxes from all methods are transformed into the same geometric representation before oriented-IoU evaluation.

## Evaluation Protocol

For dense foreground prediction, we report per-class precision, recall, and F1-score:

$$\text{Precision} = \frac{TP}{TP + FP}, \qquad \text{Recall} = \frac{TP}{TP + FN}, \tag{14}$$

and

$$F1 = \frac{2\,\text{Precision}\,\text{Recall}}{\text{Precision} + \text{Recall}}. \tag{15}$$

We additionally report foreground accuracy, macro-F1, and dense mean average precision.

For instance detection, predictions are sorted by confidence and matched one-to-one within each class and frame using oriented IoU. We report $\mathrm{mAP}_{0.50}$, $\mathrm{mAP}_{0.75}$, and

$$\mathrm{mAP}_{0.50:0.95} = \frac{1}{10C} \sum_{c=1}^{C} \sum_{\tau \in \{0.50, 0.55, \ldots, 0.95\}} \mathrm{AP}_{c,\tau} . \quad (16)$$

The unified IndoorBEV evaluation uses a confidence threshold of 0.30, an oriented-box NMS threshold of 0.20, a $3 \times 3$ local-maximum kernel, and at most 100 detections per class and frame.

Localization uses class-specific greedy one-to-one nearest-center matching within each frame. Unmatched ground-truth objects receive zero IoU. Mean oriented IoU is

$$\mathrm{mIoU} = \frac{1}{N} \sum_{i=1}^{N} \mathrm{IoU}_i, \quad (17)$$

and

$$\mathrm{Acc@IoU}_\tau = \frac{|\{i \mid \mathrm{IoU}_i \geq \tau\}|}{N}, \quad (18)$$

for $\tau \in \{0.50, 0.75\}$.

The orientation error is evaluated with the $\pi$-periodic difference

$$e_\theta = \frac{1}{2} \left| \mathrm{atan2}\left( \sin 2(\hat{\theta} - \theta), \cos 2(\hat{\theta} - \theta) \right) \right| . \quad (19)$$

Oracle analyses replace one predicted component with its ground-truth value while retaining all remaining predicted components. Center, size, and yaw oracles are used to identify the dominant source of localization error.

## Detailed Per-Class Results

### Simulated Indoor Dataset

### Real-World Indoor Dataset

## Component Ablation Results

The primary component ablations use the same direct-angle formulation as the main IndoorBEV experiments. The tables report the unified test metrics. Values are rounded to three decimal places.

### Input-Representation Ablation

On the simulated dataset, the richer representations progressively improve dense recognition and $\mathrm{mAP}_{0.50}$, with the full representation obtaining the highest dense mAP and $\mathrm{mAP}_{0.50}$. On the real-world dataset, the full representation gives the strongest aggregate instance-detection and localization results. The strict oriented-box metrics in these direct-angle ablations are also affected by the orientation behavior examined in section.“Extended Double-Angle Ablation Analysis.”.

### Fourier-Frequency Ablation

Increasing the number of Fourier frequencies does not produce a monotonic improvement. Low-order encoding with $L = 2$ provides the strongest aggregate AP on both datasets, whereas higher-frequency configurations can improve dense classification without consistently improving oriented-box localization.

### Architecture Comparison

The plain BEV-FPN has lower model-only latency. AFN provides stronger dense recognition and $\mathrm{mAP}_{0.50}$ in simulation and clear improvements across the principal real-world accuracy metrics, illustrating the accuracy-efficiency trade-off introduced by global feature fusion.

## Extended Double-Angle Ablation Analysis

The direct-angle component ablations in section."Component Ablation Results" characterize the initial IndoorBEV formulation. This section adds a parallel set of controlled ablations using the double-angle orientation representation. The original direct-angle tables are retained because they document the behavior of the initial design. The new experiments use the same datasets, fixed splits, random seed, model widths, input representations, optimization settings, checkpoint-selection rule, and evaluation protocol. Only the orientation target and associated loss are changed. All values are rounded to three decimal places, while percentage changes are computed from the unrounded results.

### Stability Across Representations and Architectures

Fig. 9 compares direct-angle and double-angle regression across overlap thresholds and controlled configurations. The double-angle representation improves every reported instance-level metric for every configuration. Averaged over the nine configurations, $\mathrm{mAP}_{0.50:0.95}$ increases from 0.253 to 0.570, $\mathrm{mAP}_{0.75}$ increases from 0.055 to 0.634, mIoU increases from 0.637 to 0.766, and $\mathrm{Acc@IoU}_{0.75}$ increases from 0.189 to 0.688. The median periodic yaw error decreases from an average of 72.447° to 0.351°.

The simulated results establish a configuration-independent effect: all nine configurations improve in mAP, $\mathrm{mAP}_{0.50}$, $\mathrm{mAP}_{0.75}$, mIoU, and $\mathrm{Acc@IoU}_{0.75}$, and all nine reduce the periodic yaw error. In the real-world data, the yaw error decreases in eight of nine configurations and $\mathrm{Acc@IoU}_{0.75}$ improves in seven of nine configurations. The less uniform changes in aggregate AP indicate that orientation periodicity is only one of several limitations under sparse and occluded real observations.

### Double-Angle Input-Representation Ablation

In simulation, the representation ablation follows the intended design progression. Adding mean height improves detection, adding height statistics improves strict localization, and Fourier height encoding provides the largest final gain. The height-statistics-plus-Fourier configuration reaches an mAP of 0.612 and mIoU of 0.782. The full representation is slightly weaker because the simulated data do not provide intensity and the additional channel is therefore identically zero.

The real-world ranking differs. Height statistics with density provide the strongest localization and the lowest median yaw error among the five representation variants, whereas the full representation gives the highest dense mAP. This result suggests that low-data real-world training can favor lower-complexity geometric features, while intensity remains useful for dense recognition.

Table 4: Per-class classification performance on the simulated indoor dataset. M1–M5 denote PillarNet, PointNet++, BEV-MAE, GroupFree3D, and IndoorBEV, respectively.

| **Class** | **M1** | | | **M2** | | | **M3** | | | **M4** | | | **M5** | | |
|---|---|---|---|---|---|---|---|---|---|---|---|---|---|---|---|
| | **P** | **R** | **F1** | **P** | **R** | **F1** | **P** | **R** | **F1** | **P** | **R** | **F1** | **P** | **R** | **F1** |
| Table | 0.65 | 0.58 | 0.61 | 0.68 | 0.60 | 0.64 | 0.55 | 0.50 | 0.52 | 0.69 | 0.62 | 0.65 | **0.98** | **0.88** | **0.93** |
| Chair | 0.62 | 0.55 | 0.58 | 0.65 | 0.58 | 0.61 | 0.52 | 0.48 | 0.50 | 0.67 | 0.60 | 0.63 | **0.98** | **0.73** | **0.84** |
| Shelf | 0.60 | 0.53 | 0.56 | 0.62 | 0.55 | 0.58 | 0.50 | 0.45 | 0.47 | 0.64 | 0.57 | 0.60 | **0.99** | **0.69** | **0.82** |
| Box | 0.58 | 0.50 | 0.54 | 0.60 | 0.52 | 0.56 | 0.48 | 0.43 | 0.45 | 0.61 | 0.54 | 0.57 | **0.99** | **0.77** | **0.87** |
| Stair | 0.63 | 0.56 | 0.59 | 0.66 | 0.59 | 0.62 | 0.53 | 0.48 | 0.50 | 0.68 | 0.61 | 0.64 | **0.99** | **0.89** | **0.94** |
| Overall Accuracy | | 0.62 | | | 0.64 | | | 0.50 | | | 0.65 | | | **0.86** | |

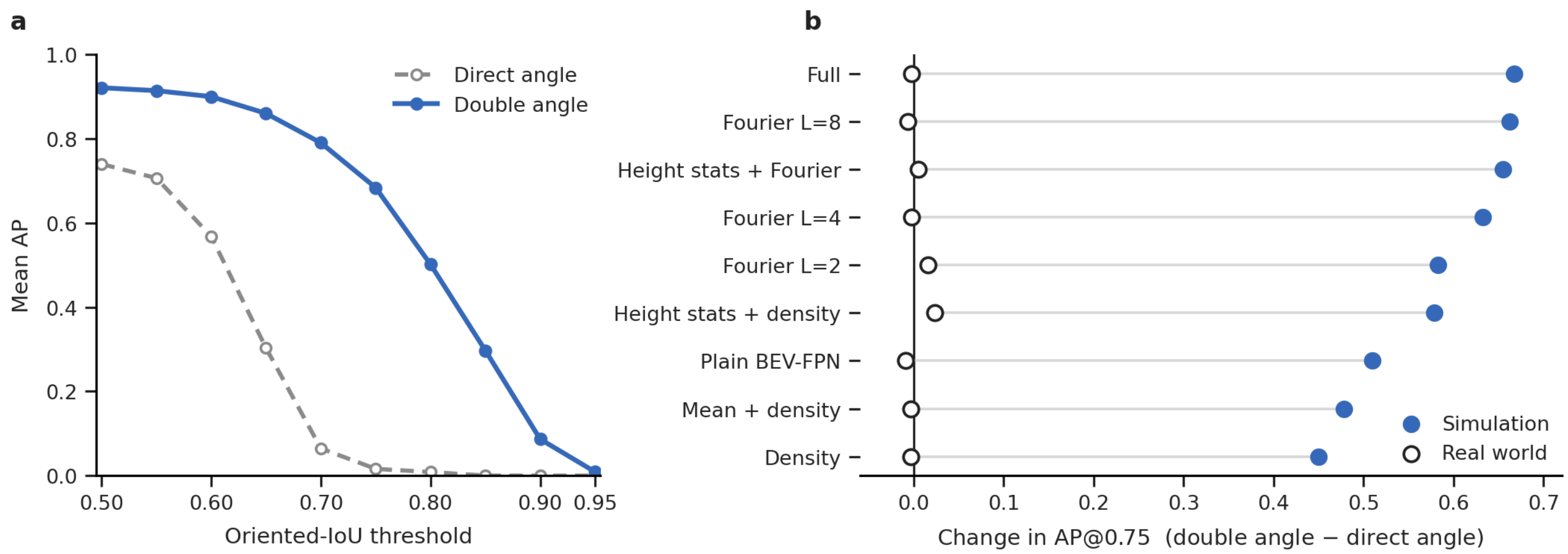

Figure 9: Stability of double-angle orientation regression across overlap thresholds, input representations, and network configurations. **a**, Mean per-class AP as a function of the oriented-IoU threshold for direct-angle and double-angle regression on the simulated indoor test set. Direct scalar-angle regression deteriorates rapidly at stricter matching thresholds, whereas double-angle regression retains substantially higher AP. **b**, Change in AP@0.75 obtained by replacing direct-angle regression with double-angle regression across the controlled configurations. Filled markers denote the simulated domain and open markers denote the real-world domain. Positive values indicate an improvement from double-angle regression.

Table 5: Per-class detection performance on the simulated indoor dataset.

| **Class** | **M1** | **M2** | **M3** | **M4** | **M5** |
|---|---|---|---|---|---|
| Table | 0.520 | 0.550 | 0.480 | 0.580 | **0.960** |
| Chair | 0.480 | 0.500 | 0.420 | 0.530 | **0.850** |
| Shelf | 0.450 | 0.470 | 0.380 | 0.500 | **0.830** |
| Box | 0.420 | 0.440 | 0.350 | 0.480 | **0.890** |
| Stair | 0.500 | 0.520 | 0.450 | 0.550 | **0.960** |
| **mAP** | 0.470 | 0.490 | 0.420 | 0.530 | **0.900** |

Table 6: Oriented-box localization performance on the simulated indoor dataset. Higher is better for mean IoU and IoU accuracy, whereas lower is better for the regression losses.

| **Metric** | **M1** | **M2** | **M3** | **M4** | **M5** |
|---|---|---|---|---|---|
| Mean IoU | 0.660 | 0.700 | 0.790 | 0.680 | **0.860** |
| Acc@IoU 0.25 | 0.890 | 0.910 | 0.920 | 0.900 | **0.990** |
| Acc@IoU 0.50 | 0.740 | 0.810 | 0.870 | 0.790 | **0.970** |
| Acc@IoU 0.75 | 0.510 | 0.580 | 0.720 | 0.560 | **0.880** |
| L1 loss | 3.040 | 2.830 | 2.370 | 2.790 | **0.880** |
| L2 loss | 66.290 | 71.500 | 66.140 | 63.460 | **3.390** |

## Double-Angle Fourier-Frequency Ablation

The simulated sweep shows that additional frequencies are not monotonically beneficial. $L = 6$ gives the best overall balance, while $L = 4$ gives the highest $\mathrm{mAP}_{0.75}$. The raw mIoU values for $L = 6$ and $L = 8$ differ by only $7 \times 10^{-6}$ and should be treated as equivalent. In the real-world dataset, $L = 2$ gives the strongest aggregate detection performance, supporting the use of lower-order encoding when the available training set is small.

Table 7: Per-class classification performance on the real-world indoor dataset.

| Class | M1 | | | M2 | | | M3 | | | M4 | | | M5 | | |
|---|---|---|---|---|---|---|---|---|---|---|---|---|---|---|---|
| | P | R | F1 | P | R | F1 | P | R | F1 | P | R | F1 | P | R | F1 |
| Couch | 0.52 | 0.45 | 0.48 | 0.50 | 0.43 | 0.46 | 0.48 | 0.40 | 0.44 | 0.50 | 0.42 | 0.46 | **1.00** | **0.73** | **0.84** |
| Table | 0.45 | 0.40 | 0.42 | 0.43 | 0.38 | 0.40 | 0.40 | 0.35 | 0.37 | 0.42 | 0.37 | 0.39 | **0.97** | **0.57** | **0.72** |
| Person | 0.42 | 0.38 | 0.40 | 0.40 | 0.35 | 0.37 | 0.38 | 0.33 | 0.35 | 0.39 | 0.34 | 0.36 | **0.98** | **0.37** | **0.53** |
| Chair | 0.40 | 0.35 | 0.37 | 0.38 | 0.33 | 0.35 | 0.35 | 0.30 | 0.32 | 0.37 | 0.32 | 0.34 | **0.96** | **0.52** | **0.68** |
| Overall Accuracy | | 0.44 | | | 0.42 | | | 0.39 | | | 0.41 | | | **0.61** | |

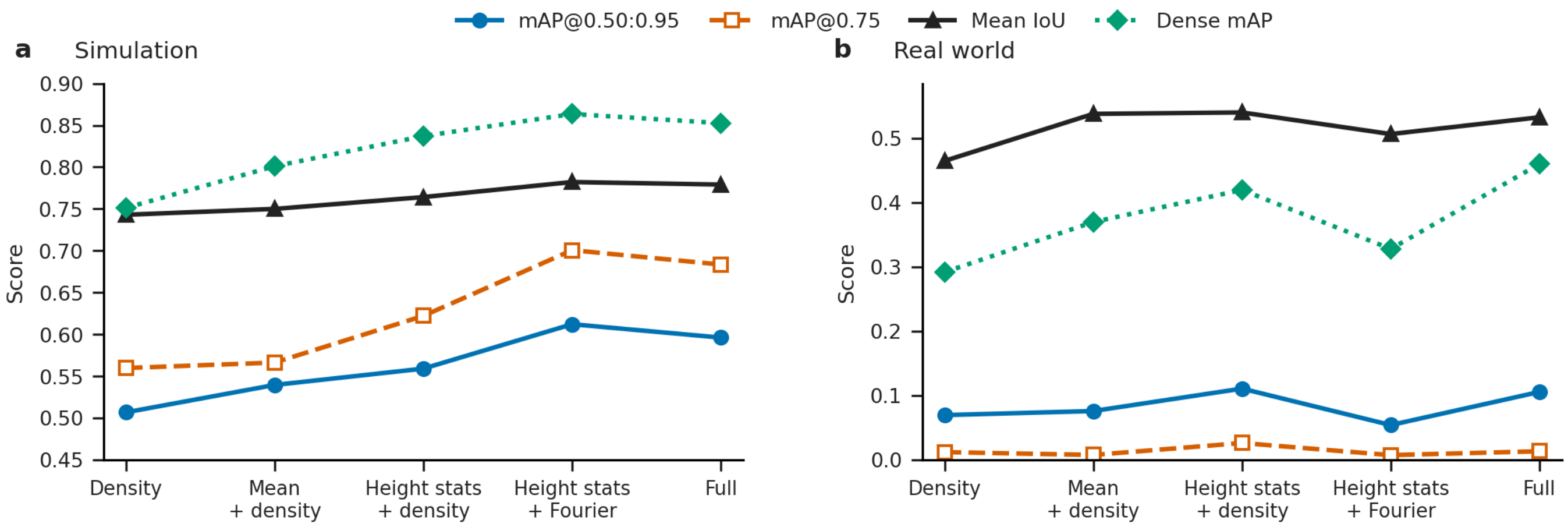

Figure 10: Input-representation ablation under double-angle orientation regression. **a**, Results on the simulated indoor dataset. Adding mean-height information, local height statistics, and Fourier height encoding progressively improves detection, localization, and dense recognition, with the height-statistics-plus-Fourier representation providing the strongest overall simulated-domain performance. **b**, Results on the real-world robot dataset. The response is less monotonic, indicating that increasing representation complexity does not uniformly improve all metrics under real sensor noise, scene sparsity, and annotation variability.

Table 8: Per-class detection performance on the real-world indoor dataset.

| Class | M1 | M2 | M3 | M4 | M5 |
|---|---|---|---|---|---|
| Couch | 0.350 | 0.380 | 0.320 | 0.360 | **0.830** |
| Table | 0.330 | 0.300 | 0.310 | 0.340 | **0.680** |
| Chair | 0.300 | 0.320 | 0.280 | 0.310 | **0.530** |
| Person | 0.320 | 0.340 | 0.290 | 0.330 | **0.470** |
| **mAP** | 0.330 | 0.340 | 0.300 | 0.340 | **0.630** |

Table 9: Oriented-box localization performance on the real-world indoor dataset. Higher is better for mean IoU and IoU accuracy, whereas lower is better for the regression losses.

| Metric | M1 | M2 | M3 | M4 | M5 |
|---|---|---|---|---|---|
| Mean IoU | **0.810** | 0.800 | 0.690 | 0.760 | 0.720 |
| Acc@IoU 0.25 | **0.990** | **0.990** | 0.940 | 0.970 | **0.990** |
| Acc@IoU 0.50 | 0.890 | 0.900 | 0.790 | 0.850 | **0.910** |
| Acc@IoU 0.75 | 0.700 | **0.720** | 0.520 | 0.620 | 0.510 |
| L1 loss | 1.280 | 1.370 | 2.160 | 1.790 | **0.800** |
| L2 loss | 12.580 | 15.250 | 29.290 | 22.370 | **2.580** |

## Double-Angle Architecture and Efficiency Analysis

The AFN advantage remains after orientation periodicity is resolved. In simulation, AFN and the plain baseline both achieve sub-degree median yaw errors, yet AFN improves $\mathrm{mAP}_{0.75}$ by 0.149 and $\mathrm{Acc@IoU}_{0.75}$ by 0.130. The improvement therefore cannot be attributed to yaw estimation alone. AFN more effectively uses global context for confidence estimation, center localization, and size regression. The real-world comparison shows the same pattern, with AFN improving mAP by approximately 0.057 and dense mAP by 0.123. The plain baseline remains approximately three times faster, defining a clear accuracy-efficiency trade-off.

## Implications for Indoor Robotic Perception

The complete ablation set separates two contributions that were confounded under direct-angle regression. First, the double-angle target provides a representation consistent with the undirected principal axis of an oriented rectangle. Second, AFN and the height-aware representation improve recognition and precise geometry after the orientation ambiguity has been removed. In simulation, yaw-oracle substitution produces almost no further improvement for any double-angle configuration, while center and size oracles remain beneficial. The remaining bottleneck is therefore geometric completion and precise extent estimation rather than orientation periodicity.

Table 10: Input-representation ablation on the simulated indoor dataset. Higher is better for all accuracy metrics, whereas lower is better for model-only latency.

| **Representation** | $\mathbf{mAP}_{0.50:0.95}$ | $\mathbf{AP}_{0.50}$ | $\mathbf{AP}_{0.75}$ | **mIoU** | **Dense mAP** | **Latency (ms)** |
|---|---|---|---|---|---|---|
| Density only | 0.255 | 0.653 | **0.110** | **0.656** | 0.773 | **18.447** |
| Mean + density | 0.253 | 0.666 | 0.088 | 0.631 | 0.801 | 18.470 |
| Height statistics + density | 0.241 | 0.714 | 0.044 | 0.631 | 0.825 | 18.542 |
| Height statistics + Fourier | **0.264** | 0.719 | 0.046 | 0.636 | 0.844 | 18.884 |
| Full | 0.241 | **0.739** | 0.016 | 0.626 | **0.862** | 18.852 |

Table 11: Input-representation ablation on the real-world indoor dataset. Higher is better for all accuracy metrics, whereas lower is better for model-only latency.

| **Representation** | $\mathbf{mAP}_{0.50:0.95}$ | $\mathbf{AP}_{0.50}$ | $\mathbf{AP}_{0.75}$ | **mIoU** | **Dense mAP** | **Latency (ms)** |
|---|---|---|---|---|---|---|
| Density only | 0.076 | 0.278 | **0.016** | 0.482 | 0.359 | 20.568 |
| Mean + density | 0.089 | 0.311 | 0.010 | 0.519 | 0.507 | 20.580 |
| Height statistics + density | 0.079 | 0.329 | 0.003 | 0.490 | 0.512 | 20.752 |
| Height statistics + Fourier | 0.084 | 0.322 | 0.002 | 0.480 | **0.547** | 20.963 |
| Full | **0.116** | **0.419** | 0.015 | **0.541** | 0.528 | **18.973** |

In the real-world dataset, double-angle encoding provides a consistent but smaller geometric benefit. The remaining gap is associated with partial visibility, sparse support, center and size uncertainty, class prediction variability, and possible ambiguity between the annotated principal axis and the axis observable from a single LiDAR frame. These results motivate future multi-sensor-assisted annotation and cross-modal supervision while preserving LiDAR-only training and inference for controlled single-modality evaluation.

## Further Study of Orientation Representation

The initial IndoorBEV formulation represents orientation using a direct yaw-regression channel trained with a Smooth L1 objective. This design is used in the primary method comparisons and component-ablation experiments.

Through subsequent experiments and further consideration of the geometric meaning of orientation in robotic perception, we found that a double-angle representation provides a more natural formulation for the evaluated task. In the BEV detection setting, the bounding box primarily describes an object axis rather than a directed heading. Consequently, orientations separated by $\pi$ represent the same rectangular geometry:

$$\theta \equiv \theta + \pi. \tag{20}$$

Direct numerical regression does not explicitly preserve this equivalence. For example, $-90^\circ$ and $+90^\circ$ describe the same undirected box axis but are treated as distant numerical targets by a conventional Smooth L1 objective.

Motivated by this observation, we further developed IndoorBEV using

$$\mathbf{v}_\theta = (\cos 2\theta, \sin 2\theta)\,, \tag{21}$$

which satisfies

$$\mathbf{v}_{\theta+\pi} = \mathbf{v}_\theta. \tag{22}$$

The orientation is decoded as

$$\hat{\theta} = \frac{1}{2}\,\mathrm{atan2}\left(\widehat{\sin 2\theta}, \widehat{\cos 2\theta}\right). \tag{23}$$

This formulation is more consistent with robotic spatial perception because it models the observable principal axis without introducing an artificial distinction between geometrically equivalent box orientations.

The predicted vector is normalized and trained using cosine distance:

$$\mathcal{L}_{\mathrm{yaw}}^{\mathrm{DA}} = \frac{\sum_{(x,y)\in\Omega_{\mathrm{reg}}}\left(1 - \hat{\mathbf{v}}_{\theta,x,y}^{\mathsf{T}}\mathbf{v}_{\theta,x,y}\right)}{|\Omega_{\mathrm{reg}}| + \epsilon}. \tag{24}$$

The double-angle regression loss is

$$\mathcal{L}_{\mathrm{reg}}^{\mathrm{DA}} = \mathcal{L}_{\mathrm{box}} + \lambda_{\mathrm{yaw}}\mathcal{L}_{\mathrm{yaw}}^{\mathrm{DA}}, \tag{25}$$

where $\lambda_{\mathrm{yaw}} = 1$.

In simulation, double-angle encoding reduces the median periodic yaw error from $58.111^\circ$ to $0.346^\circ$. The corresponding $\mathrm{mAP}_{0.75}$ increases from 0.016 to 0.683, and $\mathrm{Acc@IoU}_{0.75}$ increases from 0.060 to 0.722. The model-only latency remains effectively unchanged.

On the real-world dataset, double-angle encoding reduces the median yaw error from $40.025^\circ$ to $24.623^\circ$. The smaller improvement relative to simulation suggests that real-world orientation estimation is also affected by partial visibility, sparse point support, ambiguity in the principal object axis, and geometric annotation consistency. The aggregate AP does not increase in this run because dense classification performance is also lower. Nevertheless, the reduction in orientation error supports the geometric motivation for the revised representation.

Table 12: Fourier-frequency ablation on the simulated and real-world indoor datasets. All configurations use the height-statistics-plus-Fourier representation. The best result for each metric within each dataset is shown in bold.

| Dataset | L | $\mathbf{mAP}_{0.50:0.95}$ | $\mathbf{AP}_{0.50}$ | $\mathbf{AP}_{0.75}$ | mIoU | Dense mAP |
|---|---|---|---|---|---|---|
| Simulation | 2 | **0.273** | **0.734** | **0.083** | **0.651** | 0.843 |
| Simulation | 4 | 0.250 | 0.708 | 0.072 | 0.646 | 0.833 |
| Simulation | 6 | 0.264 | 0.719 | 0.046 | 0.636 | 0.844 |
| Simulation | 8 | 0.250 | 0.723 | 0.012 | 0.632 | **0.852** |
| Real world | 2 | **0.108** | **0.418** | 0.010 | 0.505 | 0.515 |
| Real world | 4 | 0.082 | 0.257 | 0.014 | **0.531** | 0.439 |
| Real world | 6 | 0.084 | 0.322 | 0.002 | 0.480 | **0.547** |
| Real world | 8 | 0.099 | 0.328 | **0.027** | 0.481 | 0.521 |

Table 13: Comparison between AFN and the plain BEV-FPN baseline using the full input representation and direct-angle orientation regression. Higher is better for all accuracy metrics, whereas lower is better for model-only latency. The best result within each dataset is shown in bold.

| Dataset | Architecture | $\mathbf{mAP}_{0.50:0.95}$ | $\mathbf{AP}_{0.50}$ | mIoU | Dense mAP | Latency (ms) |
|---|---|---|---|---|---|---|
| Simulation | AFN | 0.241 | **0.739** | 0.626 | **0.862** | 18.852 |
| Simulation | Plain BEV-FPN | **0.255** | 0.715 | **0.628** | 0.815 | **9.125** |
| Real world | AFN | **0.116** | **0.419** | **0.541** | **0.528** | 18.973 |
| Real world | Plain BEV-FPN | 0.072 | 0.279 | 0.512 | 0.381 | **9.887** |

For simulation, replacing the predicted yaw with ground truth produces almost no additional mIoU improvement, indicating that yaw is no longer a limiting factor. The remaining error is dominated by center and size estimation. For the real-world dataset, the size and center oracles provide larger improvements than the yaw oracle, indicating that incomplete observations and extent estimation remain important sources of error.

## Additional Embedded-System Analysis

Unless otherwise stated, embedded-system measurements are collected on an NVIDIA AGX Orin with batch size one. All evaluated methods use the same numerical precision, software environment, input samples, and timing boundaries. Model-only latency and end-to-end ROS2 latency are treated as separate measurements.

### Sequential and Pipelined Processing

The ROS2 implementation separates sensor reception, BEV construction, network inference, output publication, and diagnostics into independent callback lanes. A bounded latest-frame queue prevents stale frames from accumulating when the perception callback cannot process every received frame.

Fig. 8 compares sequential and pipelined execution across different input point-count intervals. The pipelined implementation reduces median and tail latency by overlapping or decoupling operations that do not belong to the same critical execution stage. The reduction becomes more pronounced for dense input frames.

### Capture-Time Alignment and Critical Path

Host reception time contains communication, queueing, and callback-scheduling variation and therefore does not provide a stable estimate of the original sensor-frame timestamp. IndoorBEV uses hardware capture timestamps as the origin of end-to-end latency measurement. For frame $i$,

$$T_i^{\mathrm{e2e}} = t_i^{\mathrm{pub}} - t_i^{\mathrm{cap}}, \tag{26}$$

whereas processing latency is

$$T_i^{\mathrm{proc}} = t_i^{\mathrm{pub}} - t_i^{\mathrm{start}}. \tag{27}$$

The IndoorBEV critical path consists of frame selection, ROI filtering, height-aware BEV construction, neural-network inference, output decoding, NMS, and publication. Visualization, logging, and nonessential diagnostics are executed outside the critical path whenever possible. The bounded latest-frame queue favors temporal freshness over processing every received frame and prevents unbounded queueing delay.

## Additional Open-Source Baseline Adaptation and Evaluation

This appendix describes the adaptation, training, and evaluation of two representative open-source 3D object detectors, VoteNet and 3DETR. The purpose of these experiments is to compare AFN against substantially different detection paradigms under a controlled data and evaluation protocol. VoteNet represents point-based feature extraction followed by Hough voting, whereas 3DETR represents query-based set prediction using a Transformer encoder-decoder and Hungarian matching. The original architectural components, prediction heads, matching procedures, and training objectives of both baselines were retained. Adaptations were restricted to dataset interfaces, coordinate conventions, ROCm compatibility, and the deterministic projection of native 3D box predictions into the common bird's-eye-view evaluation space.

Table 14: Average effect of orientation representation across the nine controlled configurations for each dataset. Higher is better for all detection and localization metrics, whereas lower is better for the median periodic yaw error. The best result within each dataset is shown in bold.

| Dataset | Orientation representation | $\mathbf{mAP}_{0.50:0.95}$ | $\mathbf{AP}_{0.50}$ | $\mathbf{AP}_{0.75}$ | mIoU | Acc@IoU 0.75 | Median yaw error |
|---|---|---|---|---|---|---|---|
| Simulation | Direct angle | 0.253 | 0.708 | 0.055 | 0.637 | 0.189 | 72.447° |
| Simulation | Double angle | **0.570** | **0.912** | **0.634** | **0.766** | **0.688** | **0.351°** |
| Real world | Direct angle | **0.090** | **0.327** | 0.012 | 0.504 | 0.075 | 35.491° |
| Real world | Double angle | 0.087 | 0.312 | **0.014** | **0.510** | **0.095** | **28.391°** |

Table 15: Double-angle input-representation ablation on the simulated indoor test set. Higher is better for all detection, localization, and dense-recognition metrics, whereas lower is better for model-only latency. The best result for each metric is shown in bold.

| Representation | $\mathbf{mAP}_{0.50:0.95}$ | $\mathbf{AP}_{0.50}$ | $\mathbf{AP}_{0.75}$ | mIoU | Acc@IoU 0.50 | Acc@IoU 0.75 | Dense mAP | Latency (ms) |
|---|---|---|---|---|---|---|---|---|
| Density only | 0.507 | 0.825 | 0.560 | 0.743 | 0.935 | 0.653 | 0.751 | **18.373** |
| Mean + density | 0.540 | 0.892 | 0.566 | 0.750 | 0.943 | 0.647 | 0.801 | 18.391 |
| Height statistics + density | 0.559 | 0.925 | 0.622 | 0.764 | 0.953 | 0.693 | 0.837 | 18.484 |
| Height statistics + Fourier | **0.612** | **0.940** | **0.701** | **0.782** | **0.961** | **0.733** | **0.864** | 18.822 |
| Full | 0.596 | 0.920 | 0.683 | 0.779 | 0.957 | 0.722 | 0.853 | 18.910 |

## Compared Methods

The comparison includes the following methods:

- **AFN**: the proposed height-aware BEV detector using the full six-channel input representation and double-angle orientation regression.
- **VoteNet**: a point-based 3D detector that extracts PointNet++ features, predicts votes toward object centers, aggregates the votes, and generates object proposals.
- **3DETR**: a Transformer-based detector that constructs non-parametric object queries from point-cloud features and performs set-based box prediction using Hungarian matching.

AFN directly predicts oriented BEV boxes parameterized by $(x, y, l, w, \theta)$. VoteNet and 3DETR natively predict full 3D boxes. For the common evaluation, the native 3D predictions are projected to BEV by retaining the horizontal center, horizontal box dimensions, orientation, semantic class, and confidence score. No prediction head or training loss is removed from either open-source baseline.

## Datasets and Class Definitions

The same simulation and real-world splits used by AFN are also used for VoteNet and 3DETR. Each point-cloud frame is sampled to 24,000 XYZ points. RGB information and external pretraining are not used. The simulation dataset contains five foreground classes:

$$\mathcal{C}_{\mathrm{sim}} = \{\mathrm{Table}, \mathrm{Chair}, \mathrm{Shelf}, \mathrm{Box}, \mathrm{Stair}\}. \tag{28}$$

The real-world dataset contains four foreground classes:

$$\mathcal{C}_{\mathrm{real}} = \{\mathrm{Couch}, \mathrm{Table}, \mathrm{Person}, \mathrm{Chair}\}. \tag{29}$$

The simulation split contains 2,029 training samples and 255 test samples. The real-world split contains 416 training samples, 52 validation samples, and 52 test samples. The same split files are used for all compared methods.

## Point Sampling

Each point-cloud frame is represented by 24,000 XYZ points. Let the original number of finite points in a frame be $N$. If $N \geq 24{,}000$, points are sampled without replacement. If $N < 24{,}000$, points are sampled with replacement. This procedure produces an input tensor of shape

$$\mathbf{P} \in \mathbb{R}^{24{,}000\times 3}. \tag{30}$$

The point budget is identical across VoteNet and 3DETR. AFN consumes the same observations after deterministic BEV rasterization.

## Coordinate Conventions

**Simulation Labels** Simulation labels are already represented in the common LiDAR coordinate system. The object center, dimensions, and heading are read as

$$\mathbf{c} = (x, y, z), \qquad \mathbf{s} = (l, w, h), \qquad \theta = \theta_{\mathrm{label}}. \tag{31}$$

**Real-World Labels** The real-world labels follow the KITTI camera-coordinate convention. The camera-coordinate location is denoted by $(x_{\mathrm{cam}}, y_{\mathrm{cam}}, z_{\mathrm{cam}})$, and the camera heading is denoted by $r_y$. The horizontal conversion used by all methods is

$$x = z_{\mathrm{cam}}, \qquad y = -x_{\mathrm{cam}}, \qquad \theta = -r_y - \frac{\pi}{2}. \tag{32}$$

Table 16: Double-angle input-representation ablation on the real-world indoor test set. Higher is better for all detection, localization, and dense-recognition metrics, whereas lower is better for the median periodic yaw error. The best result for each metric is shown in bold.

| Representation | $\mathbf{mAP}_{0.50:0.95}$ | $\mathbf{AP}_{0.50}$ | $\mathbf{AP}_{0.75}$ | mIoU | Acc@IoU 0.50 | Acc@IoU 0.75 | Dense mAP | Yaw error |
|---|---|---|---|---|---|---|---|---|
| Density only | 0.069 | 0.244 | 0.012 | 0.465 | 0.605 | **0.113** | 0.292 | 30.908° |
| Mean + density | 0.076 | 0.281 | 0.008 | 0.538 | 0.692 | 0.082 | 0.370 | 32.341° |
| Height statistics + density | **0.111** | **0.410** | **0.026** | **0.540** | **0.749** | 0.087 | 0.419 | **21.623°** |
| Height statistics + Fourier | 0.054 | 0.205 | 0.007 | 0.506 | 0.651 | 0.092 | 0.328 | 32.969° |
| Full | 0.106 | 0.395 | 0.013 | 0.532 | 0.672 | 0.103 | **0.460** | 24.623° |

Table 17: Double-angle Fourier-frequency ablation on the simulated and real-world indoor datasets. All configurations use height statistics with Fourier encoding. The best result for each metric within each dataset is shown in bold.

| Dataset | L | $\mathbf{mAP}_{0.50:0.95}$ | $\mathbf{AP}_{0.50}$ | $\mathbf{AP}_{0.75}$ | mIoU | Acc@IoU 0.75 | Dense mAP |
|---|---|---|---|---|---|---|---|
| Simulation | 2 | 0.590 | 0.930 | 0.665 | 0.777 | 0.713 | 0.848 |
| Simulation | 4 | 0.608 | **0.940** | **0.704** | 0.775 | 0.714 | 0.860 |
| Simulation | 6 | **0.612** | **0.940** | 0.701 | **0.782** | **0.733** | **0.864** |
| Simulation | 8 | 0.595 | 0.938 | 0.674 | **0.782** | 0.722 | 0.848 |
| Real world | 2 | **0.135** | **0.450** | **0.026** | **0.525** | 0.087 | **0.527** |
| Real world | 4 | 0.091 | 0.309 | 0.011 | 0.498 | 0.097 | 0.472 |
| Real world | 6 | 0.054 | 0.205 | 0.007 | 0.506 | 0.092 | 0.328 |
| Real world | 8 | 0.096 | 0.304 | 0.020 | 0.504 | **0.138** | 0.427 |

For VoteNet and 3DETR, the full 3D geometric center is required. Because the KITTI location convention represents the bottom center of the box, the vertical geometric center is computed as

$$z = -y_{\text{cam}} + \frac{h}{2}. \tag{33}$$

The resulting box representation is

$$\mathbf{b} = (x, y, z, l, w, h, \theta). \tag{34}$$

The vertical-center convention in Eq. (33) was selected through a point-occupancy audit over 200 training frames and 1,420 labeled objects. Three candidate vertical-center definitions were compared. The results are shown in Table 21. The corrected convention increases the mean number of enclosed points by approximately $7.753\%$ relative to the original interpretation and eliminates empty labeled boxes in the audited subset. AFN does not require retraining after this correction because its dense BEV target construction uses only $x$, $y$, $l$, $w$, and $\theta$. The vertical center and height are not used by the AFN training objective. VoteNet and 3DETR were retrained because both methods regress full 3D boxes.

## VoteNet Adaptation

The VoteNet dataset adapter provides the original model with the following fields:

- sampled XYZ point clouds;
- object-center labels;
- heading-class and heading-residual labels;
- size-class and size-residual labels;
- semantic-class labels;
- valid-object masks;
- point-level voting labels and voting masks.

The class-specific mean box sizes are computed from the training split only. For the real-world dataset, the resulting mean sizes in $(l, w, h)$ order are

$$\begin{aligned} \boldsymbol{\mu}_{\text{Couch}} &= (1.959,\ 1.043,\ 1.010), \\ \boldsymbol{\mu}_{\text{Table}} &= (0.983,\ 0.882,\ 0.760), \\ \boldsymbol{\mu}_{\text{Person}} &= (0.736,\ 0.811,\ 1.771), \\ \boldsymbol{\mu}_{\text{Chair}} &= (0.671,\ 0.714,\ 0.822). \end{aligned} \tag{35}$$

The corresponding training-box counts are 497, 507, 732, and 1,317, respectively. The original VoteNet components are retained:

- PointNet++ backbone;
- voting module;
- vote aggregation;
- objectness prediction;
- center regression;
- heading classification and residual regression;
- size-cluster classification and residual regression;
- semantic-class prediction;
- original VoteNet losses.

The evaluated VoteNet model contains 948,217 parameters on the simulation dataset and 947,572 parameters on the real-world dataset. The difference is caused by the number of semantic output classes.

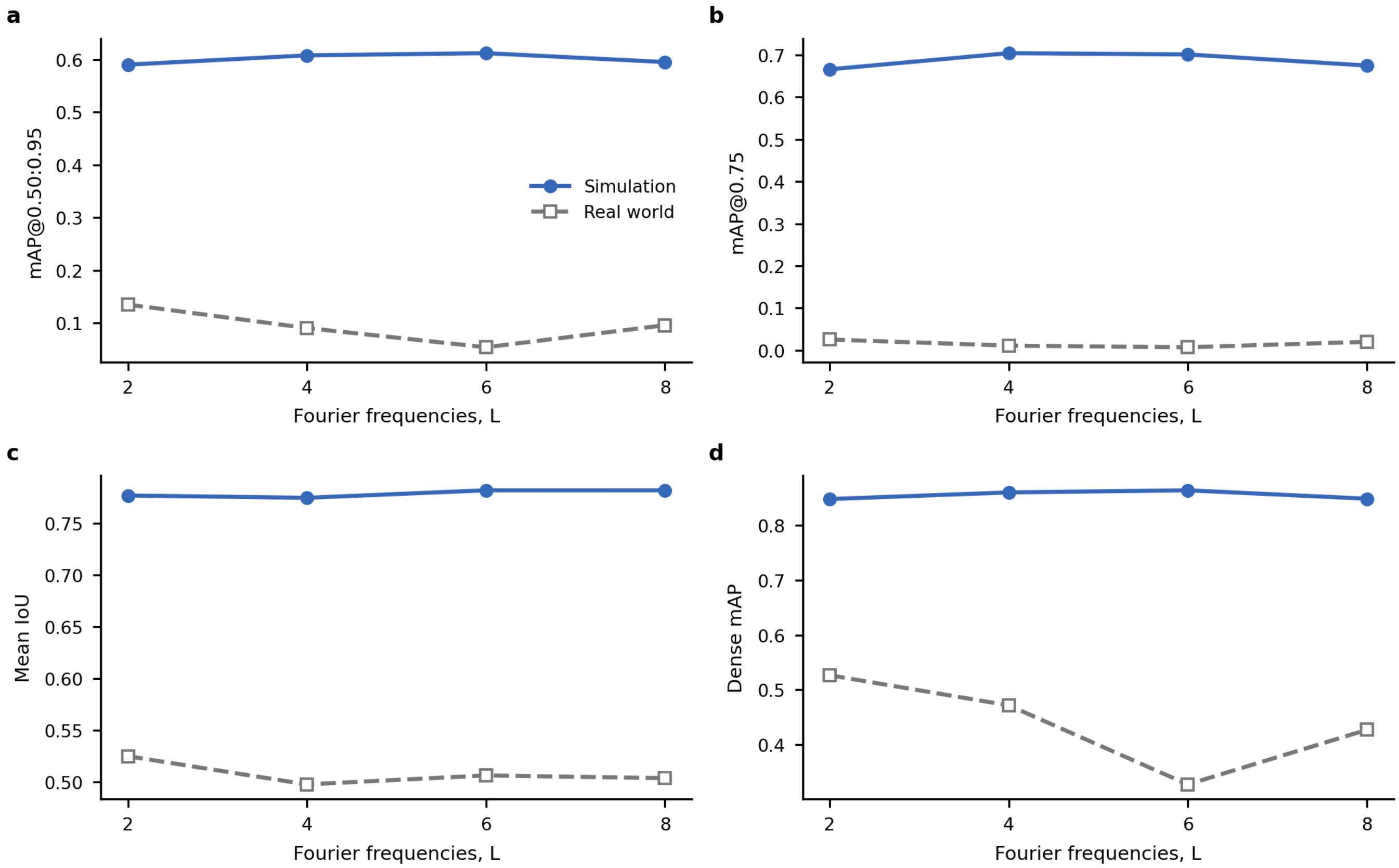

Figure 11: Effect of the Fourier height-encoding frequency under double-angle orientation regression. Results are reported for $L \in \{2, 4, 6, 8\}$ on the simulated indoor and real-world robot datasets. **a**, mAP@0.50:0.95. **b**, AP@0.75. **c**, Mean oriented IoU. **d**, Dense mAP. The simulated-domain response is comparatively stable, with the intermediate-frequency configurations providing the strongest overall balance across metrics. The real-world response is more variable and metric dependent: lower-frequency encoding yields the strongest aggregate detection and dense-recognition results, whereas mean IoU varies less substantially across the evaluated frequencies.

## 3DETR Adaptation

The 3DETR adapter returns the following fields:

- `point_clouds`;
- `gt_box_corners`;
- `gt_box_centers`;
- `gt_box_centers_normalized`;
- `gt_box_sizes`;
- `gt_box_sizes_normalized`;
- `gt_box_angles`;
- `gt_angle_class_label`;
- `gt_angle_residual_label`;
- `gt_box_sem_cls_label`;
- `gt_box_present`;
- point-cloud minimum and maximum bounds.

Box centers are normalized to the unit cube using the sampled point-cloud bounds. Box sizes are normalized elementwise by the point-cloud spatial extent. The original 12-bin heading representation is retained. The evaluated 3DETR configuration uses:

- 256 object queries;
- a 3-layer Transformer encoder;
- an 8-layer Transformer decoder;
- encoder and decoder dimensions of 256;
- four attention heads;
- 2,048 pre-encoder points;
- seven auxiliary decoder outputs;
- the original Hungarian matcher;
- the original center, size, angle, semantic, and overlap losses.

The simulation and real-world 3DETR models contain 7,308,004 and 7,307,747 parameters, respectively.

## ROCm Compatibility

The open-source implementations were adapted to PyTorch ROCm for execution on AMD Instinct MI250X GPUs. These changes were implementation-level compatibility changes and did not alter the algorithms. For the PointNet++ operators, the following changes were applied:

Table 18: Double-angle architecture comparison using the full input representation.

| Dataset | Model | $\text{mAP}_{0.50:0.95}$ | $\text{mAP}_{0.50}$ | $\text{mAP}_{0.75}$ | mIoU | Acc@0.75 | Dense mAP | Latency (ms) |
|---|---|---|---|---|---|---|---|---|
| Simulation | AFN | **0.596** | **0.920** | **0.683** | **0.779** | **0.722** | **0.853** | 18.910 |
| Simulation | Plain BEV-FPN | 0.526 | 0.901 | 0.534 | 0.739 | 0.592 | 0.800 | **6.004** |
| Real world | AFN | **0.106** | **0.395** | **0.013** | **0.532** | **0.103** | **0.460** | 18.851 |
| Real world | Plain BEV-FPN | 0.049 | 0.215 | 0.002 | 0.482 | 0.056 | 0.337 | **6.023** |

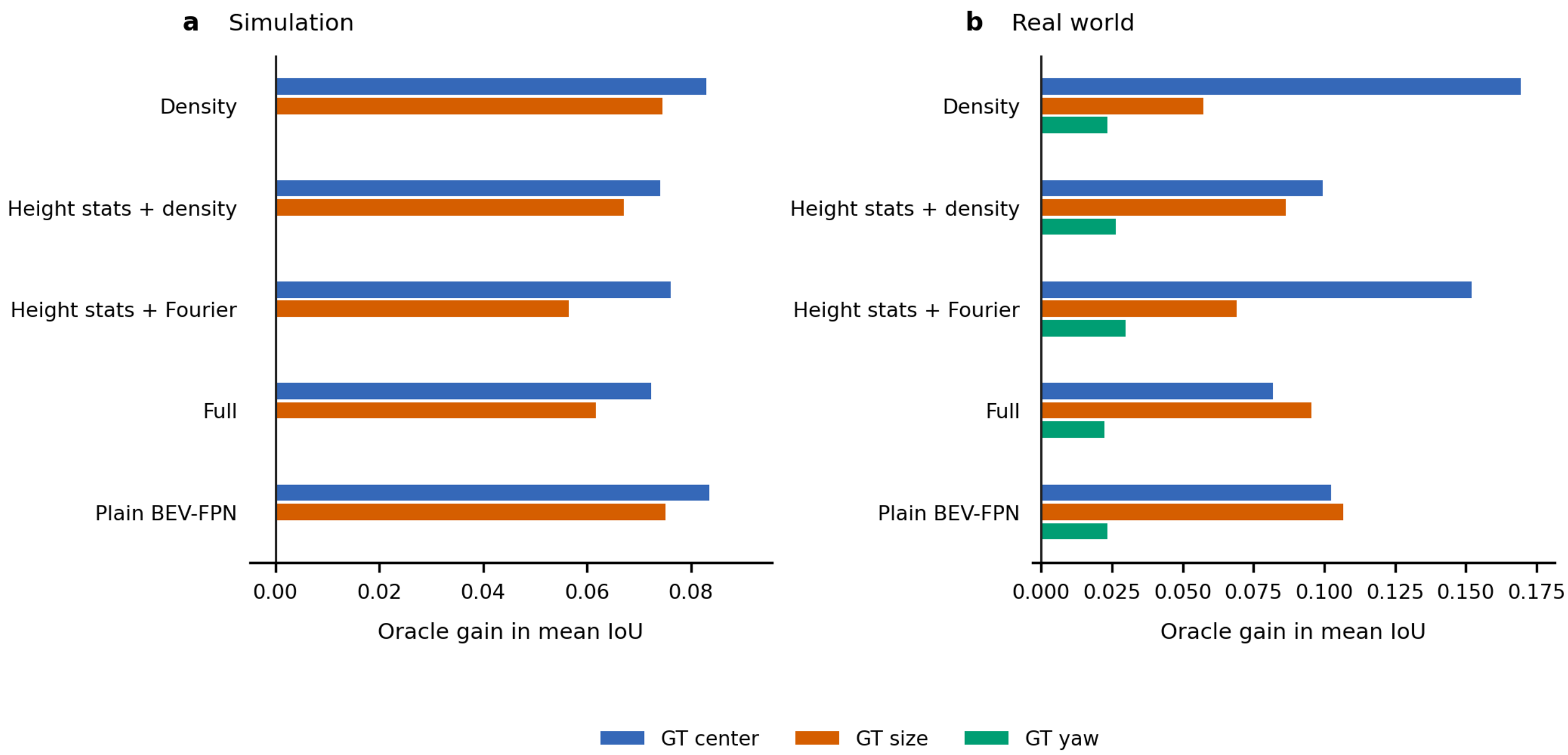

Figure 12: Oracle decomposition of the remaining oriented-box localization error under double-angle regression. Each bar reports the increase in mean oriented IoU obtained by replacing one predicted box component with its ground-truth value while retaining the remaining predicted components. **a**, Simulated indoor dataset. Ground-truth center and size yield substantial improvements, whereas replacing the predicted yaw provides negligible additional benefit. **b**, Real-world robot dataset. Center and size remain the dominant localization bottlenecks, while the yaw oracle provides a smaller but non-zero improvement.

- the original extension was built through PyTorch ROCm and HIP;
- the target GPU architecture was set to `gfx90a`;
- absolute include paths were used during extension compilation;
- removed `AT_CHECK` calls were replaced with `TORCH_CHECK`;
- the original sampling, grouping, interpolation, and gradient operations were retained.

The original 3DETR Transformer code also contained in-place Dropout operations that conflicted with the version-counter checks in PyTorch 2.10 autograd. The following compatibility-only change was applied:

$$\texttt{inplace=True} \longrightarrow \texttt{inplace=False}. \quad (36)$$

This change preserves the Dropout probability, random masking distribution, network structure, parameter count, and optimization objective. The optional Cython implementation of 3D box intersection caused native segmentation faults for specific real-world box configurations. The extension was therefore disabled, and the original Python fallback was used. This affects execution speed but does not change the mathematical definition of the overlap computation.

## Controlled Training Protocol

The controlled training settings are summarized in Table 22. The controlled setting is intended to isolate architectural differences. It is not presented as an unrestricted, method-specific hyperparameter search. Each method retains its native model structure and loss formulation, while the data, point budget, split, seed, and training budget are controlled.

## Evaluation Protocol

**Native 3D Evaluation** VoteNet and 3DETR are first evaluated with their original full-3D box evaluation procedures. These results are retained as supplementary diagnostics. AFN is not included in the full-3D evaluation because AFN does not predict the vertical center or object height.

Table 19: Further evaluation of orientation representations in IndoorBEV. The direct-angle formulation corresponds to the initial design, while the double-angle formulation incorporates the geometric $\pi$-periodicity of oriented bounding boxes.

| Dataset | Yaw representation | $\mathbf{mAP}_{0.50:0.95}$ | $\mathbf{mAP}_{0.50}$ | $\mathbf{mAP}_{0.75}$ | mIoU | Acc@0.75 | Median yaw error |
|---|---|---|---|---|---|---|---|
| Simulation | Direct angle | 0.241 | 0.739 | 0.016 | 0.626 | 0.060 | 58.111° |
| Simulation | Double angle | **0.596** | **0.920** | **0.683** | **0.779** | **0.722** | **0.346°** |
| Real world | Direct angle | **0.116** | **0.419** | **0.015** | **0.541** | 0.092 | 40.025° |
| Real world | Double angle | 0.106 | 0.395 | 0.013 | 0.532 | **0.103** | **24.623°** |

Table 20: Oracle mIoU analysis for the double-angle variants. Each oracle replaces one predicted box component with its ground-truth value.

| Dataset | Original | GT center | GT size | GT yaw |
|---|---|---|---|---|
| Simulation | 0.779 | 0.852 | 0.841 | 0.779 |
| Real world | 0.532 | 0.614 | 0.628 | 0.555 |

Table 21: Point-occupancy audit of candidate vertical-center conventions on the real-world dataset. Higher point counts and a lower empty-box fraction indicate better alignment between labeled boxes and the point cloud.

| Vertical center | Mean points | Median points | Empty fraction |
|---|---|---|---|
| $z = y_{\text{cam}}$ | 296.745 | 153.000 | 0.009 |
| $z = -y_{\text{cam}}$ | 182.249 | 95.000 | 0.001 |
| $z = -y_{\text{cam}} + h/2$ | **319.753** | **176.500** | **0.000** |

**Unified Oriented-BEV Evaluation** The primary comparison uses a common oriented-BEV protocol. Native VoteNet and 3DETR predictions are projected as

$$(x, y, z, l, w, h, \theta) \quad \longrightarrow \quad (x, y, l, w, \theta). \tag{37}$$

This projection is deterministic and does not retrain or alter either baseline. All methods are evaluated using the verified AFN evaluation implementation. A prediction is represented by

$$\hat{\mathbf{b}} = (\hat{x}, \hat{y}, \hat{l}, \hat{w}, \hat{\theta}), \tag{38}$$

together with a semantic class and confidence score. Oriented IoU is computed using rotated rectangle intersection. For each class, predictions from all test frames are sorted by confidence. A prediction is considered a true positive if the following conditions hold:

1. the semantic class matches;
2. the prediction and ground-truth box belong to the same frame;
3. the oriented IoU is at least the evaluation threshold;
4. the ground-truth object has not been matched by a higher-scoring prediction.

Average precision is calculated using 101-point interpolated precision:

$$\text{AP} = \frac{1}{101} \sum_{r \in \{0, 0.01, \ldots, 1\}} \max_{\tilde{r} \geq r} p(\tilde{r}). \tag{39}$$

Table 22: Controlled training protocol for the open-source baselines.

| Setting | Value |
|---|---|
| Input points per frame | 24,000 |
| Batch size | 4 |
| Maximum epochs | 500 |
| Learning rate | $1 \times 10^{-4}$ |
| Weight decay | $1 \times 10^{-4}$ |
| Random seed | 42 |
| Learning-rate step interval | 30 epochs |
| Learning-rate decay factor | 0.1 |
| Early-stopping patience | 45 epochs |
| Early-stopping warm-up | 30 epochs |
| Minimum validation improvement | $1 \times 10^{-4}$ |
| External pretraining | None |
| RGB input | Disabled |

The principal metric is the class mean of AP averaged across thresholds

$$\mathcal{T} = \{0.50, 0.55, \ldots, 0.95\}. \tag{40}$$

The evaluation therefore reports

$$\text{mAP}_{0.50:0.95}, \qquad \text{mAP}_{0.50}, \qquad \text{mAP}_{0.75}. \tag{41}$$

AFN foreground classes use one-based class IDs because zero denotes background. VoteNet and 3DETR use zero-based semantic IDs. The external prediction wrapper maps all external class names to the corresponding AFN foreground IDs before evaluation.

## Evaluator Regression Validation

Before evaluating external methods, the prediction-import wrapper was validated by re-evaluating AFN's saved predictions. The wrapper directly calls the original AFN functions for rotated IoU, score-sorted matching, and interpolated AP. The regression results are shown in Table 23. The imported-prediction results exactly reproduce the original AFN evaluation to six decimal places. This exact regression is important because an independently implemented evaluator produced substantial metric deviations. All final cross-method results therefore use the original, regression-validated AFN evaluation functions.

## Unified BEV Results

Table 24 reports the final unified BEV results. The unrounded results used to construct Table 24 are provided in Table 25.

Table 23: Regression validation of the external-prediction evaluation wrapper.

| Dataset | Evaluation | $\mathbf{mAP}_{.5:.95}$ | $\mathbf{mAP}_{.5}$ | $\mathbf{mAP}_{.75}$ |
|---|---|---|---|---|
| Simulation | AFN | .596 | .920 | .683 |
| Simulation | Imported | .596 | .920 | .683 |
| Real world | AFN | .105 | .395 | .013 |
| Real world | Imported | .106 | .395 | 0.013 |

## Analysis of the Open-Source Baselines

**Simulation Dataset** On the simulation dataset, AFN obtains the highest result for every reported BEV detection metric. Relative to 3DETR, AFN improves $\mathrm{mAP}_{0.50:0.95}$ from 0.491263 to 0.596127, $\mathrm{mAP}_{0.50}$ from 0.777415 to 0.920457, and $\mathrm{mAP}_{0.75}$ from 0.508367 to 0.683484. Using the unrounded values, the corresponding relative improvements are

$$\frac{0.596127 - 0.491263}{0.491263} \times 100 = 21.346\%, \tag{42}$$

$$\frac{0.920457 - 0.777415}{0.777415} \times 100 = 18.399\%, \tag{43}$$

and

$$\frac{0.683484 - 0.508367}{0.508367} \times 100 = 34.447\%. \tag{44}$$

The largest relative improvement occurs at IoU 0.75. This result indicates that AFN's advantage is not limited to object discovery at a permissive IoU threshold. AFN also produces more accurate horizontal centers, dimensions, and orientations. VoteNet obtains substantially lower simulation performance. AFN improves $\mathrm{mAP}_{0.50}$ from 0.210296 to 0.920457 and $\mathrm{mAP}_{0.75}$ from 0.027800 to 0.683484. The result suggests that the explicit BEV representation offers a strong task-specific inductive bias for the current indoor scenes.

**Real-World Dataset** On the real-world dataset, AFN achieves substantially higher overall mAP and AP at IoU 0.50 than both open-source baselines. AFN obtains $\mathrm{mAP}_{0.50:0.95} = 0.105560$, compared with 0.037915 for 3DETR. At IoU 0.50, AFN reaches 0.394526, compared with 0.115288 for 3DETR. Using the unrounded values, AFN is approximately 2.784 times higher than 3DETR in $\mathrm{mAP}_{0.50:0.95}$ and approximately 3.422 times higher at IoU 0.50. At the stricter IoU threshold of 0.75, 3DETR obtains 0.024033, whereas AFN obtains 0.013288. This result indicates that a subset of successful 3DETR predictions is localized with high precision. The result does not overturn the overall comparison because AFN retains substantially higher performance when averaged over all IoU thresholds and at IoU 0.50. VoteNet exhibits limited real-world generalization under the controlled training protocol. This behavior remains after correcting the vertical box-center convention, retraining the model, retaining low-confidence candidates, and projecting the native 3D predictions into BEV. The result suggests that the dominant errors involve horizontal localization, box dimensions, orientation, semantic confidence, or a combination of these factors.

## Accuracy and Parameter Efficiency

AFN contains approximately 0.769 million parameters, compared with 0.948 million for VoteNet and 7.308 million for 3DETR. Relative to VoteNet, the approximate parameter reduction is

$$\left(1 - \frac{0.769}{0.948}\right) \times 100 \approx 18.9\%. \tag{45}$$

Relative to 3DETR, the approximate parameter reduction is

$$\left(1 - \frac{0.769}{7.308}\right) \times 100 \approx 89.5\%. \tag{46}$$

Equivalently, 3DETR contains approximately 9.5 times as many parameters as AFN. Despite its smaller capacity, AFN obtains the highest overall simulation performance and the highest real-world $\mathrm{mAP}_{0.50:0.95}$ and $\mathrm{mAP}_{0.50}$. These results show that AFN's performance does not result from increased model capacity. Instead, the compact architecture converts a task-specific BEV representation into a stronger accuracy-parameter trade-off than either point voting or general query-based Transformer detection.

## Runtime and Memory

Table 26 summarizes the available runtime measurements.

The simulation latency measurements for VoteNet and 3DETR represent batch-level latency and should not be interpreted as batch-size-one latency. Their approximate amortized latencies are

$$\frac{88.947}{3} = 29.649 \text{ ms/sample} \tag{47}$$

for VoteNet and

$$\frac{105.932}{3} = 35.311 \text{ ms/sample} \tag{48}$$

for 3DETR. These values are included only as throughput-oriented diagnostics because the batch configuration differs from that of AFN. The real-world AFN and 3DETR measurements both use batch size one and are therefore directly comparable:

$$\frac{85.959}{18.901} = 4.548. \tag{49}$$

Thus, 3DETR has approximately 4.548 times the model-only latency of AFN on the real-world test set. Equivalently, AFN reduces model-only latency by approximately

$$\left(1 - \frac{18.901}{85.959}\right) \times 100 = 78.011\%. \tag{50}$$

## Interpretation and Limitations

Several considerations are important when interpreting the results. First, the unified comparison evaluates oriented BEV detection because this is the common output space supported by all methods. VoteNet and 3DETR also predict vertical position and object height, but these dimensions are not used in the primary comparison. Their native full-3D results remain useful as supplementary diagnostics. Second, the experiment uses a controlled training protocol rather than independent method-specific hyperparameter searches. This design isolates architectural differences but does not claim that every baseline is evaluated under its individually optimal configuration. Third, the real-world dataset is considerably smaller than the simulation dataset. The point-voting and Transformer baselines may be more sensitive to the limited training-set size than AFN's explicit BEV representation. The real-world conclusions should therefore be interpreted as performance under the stated controlled and data-limited setting. Fourth, execution constraints on the MI250X required some baseline evaluations to use batch sizes greater than one. Detection metrics are computed per sample and remain valid, but latency values from different batch sizes are not treated as strictly equivalent. Finally, the real-world AP at IoU 0.75 remains low for all methods. This suggests that highly accurate box geometry remains challenging under real-world sensing noise, occlusion, and limited supervision. Although AFN provides the strongest overall accuracy-efficiency trade-off, improving strict-IoU localization on real data remains an important direction for future work.

Table 24: Comparison with representative open-source detectors under the same oriented-BEV evaluation protocol. All methods use the same test splits and 24,000 input points.

| **Dataset** | **Model** | $\mathbf{mAP}_{0.50:0.95}$ | $\mathbf{mAP}_{0.50}$ | $\mathbf{mAP}_{0.75}$ | **Parameters (M)** |
|---|---|---|---|---|---|
| Simulation | AFN | **0.596** | **0.920** | **0.683** | **0.769** |
| Simulation | VoteNet | 0.073 | 0.210 | 0.028 | 0.948 |
| Simulation | 3DETR | 0.491 | 0.777 | 0.508 | 7.308 |
| Real world | AFN | **0.106** | **0.395** | 0.013 | **0.769** |
| Real world | VoteNet | 0.000 | 0.002 | 0.000 | 0.948 |
| Real world | 3DETR | 0.038 | 0.115 | **0.024** | 7.308 |

Table 25: Unrounded unified oriented-BEV results.

| **Dataset** | **Model** | $\mathbf{mAP}_{0.50:0.95}$ | $\mathbf{mAP}_{0.50}$ | $\mathbf{mAP}_{0.75}$ |
|---|---|---|---|---|
| Simulation | AFN | 0.596127 | 0.920457 | 0.683484 |
| Simulation | VoteNet | 0.073012 | 0.210296 | 0.027800 |
| Simulation | 3DETR | 0.491263 | 0.777415 | 0.508367 |
| Real world | AFN | 0.105560 | 0.394526 | 0.013288 |
| Real world | VoteNet | 0.000328 | 0.001660 | 0.000009 |
| Real world | 3DETR | 0.037915 | 0.115288 | 0.024033 |

Table 26: Model-only runtime and peak allocated GPU memory. Batch size is reported because MI250X MIOpen exhibited instability for batch-size-one PointNet++ execution in some baseline configurations.

| **Dataset** | **Model** | **Batch size** | **Med latency (ms/batch)** | **Peak memory (MiB)** |
|---|---|---|---|---|
| Simulation | AFN | 1 | 18.876 | - |
| Simulation | VoteNet | 3 | 88.947 | 200.073 |
| Simulation | 3DETR | 3 | 105.932 | 1188.829 |
| Real world | AFN | 1 | 18.901 | - |
| Real world | VoteNet | 2 | 94.776 | 396.444 |
| Real world | 3DETR | 1 | 85.959 | 427.512 |

## Reproducibility Artifacts

The following artifacts are retained for each baseline experiment:

- the resolved training configuration;
- epoch-level training and validation histories;
- best and latest checkpoints;
- test-set prediction files;
- native evaluation results;
- unified AFN-protocol evaluation results;
- parameter counts;
- latency and peak-memory measurements;
- dataset adaptation records;
- compatibility patches.

. The real-world VoteNet and 3DETR results use the checkpoints retrained after the vertical-center correction in Eq. (33). The corresponding run directories and unique job identifiers are retained with the generated prediction and evaluation files.

## Summary

The open-source baseline comparison supports three main conclusions. First, AFN substantially outperforms VoteNet under the same oriented-BEV evaluation protocol on both datasets. This advantage is achieved while AFN uses approximately 19% fewer parameters. Second, AFN outperforms the substantially larger 3DETR model in all three simulation AP metrics and in the principal real-world metrics, $\mathrm{mAP}_{0.50:0.95}$ and $\mathrm{mAP}_{0.50}$. 3DETR retains a modest real-world advantage at IoU 0.75. Third, AFN provides the strongest overall accuracy-efficiency trade-off among the evaluated methods. The result indicates that the proposed height-aware BEV representation and compact detection architecture provide a more effective inductive bias for the evaluated indoor detection task than either point-based voting or a substantially larger generic Transformer detector.

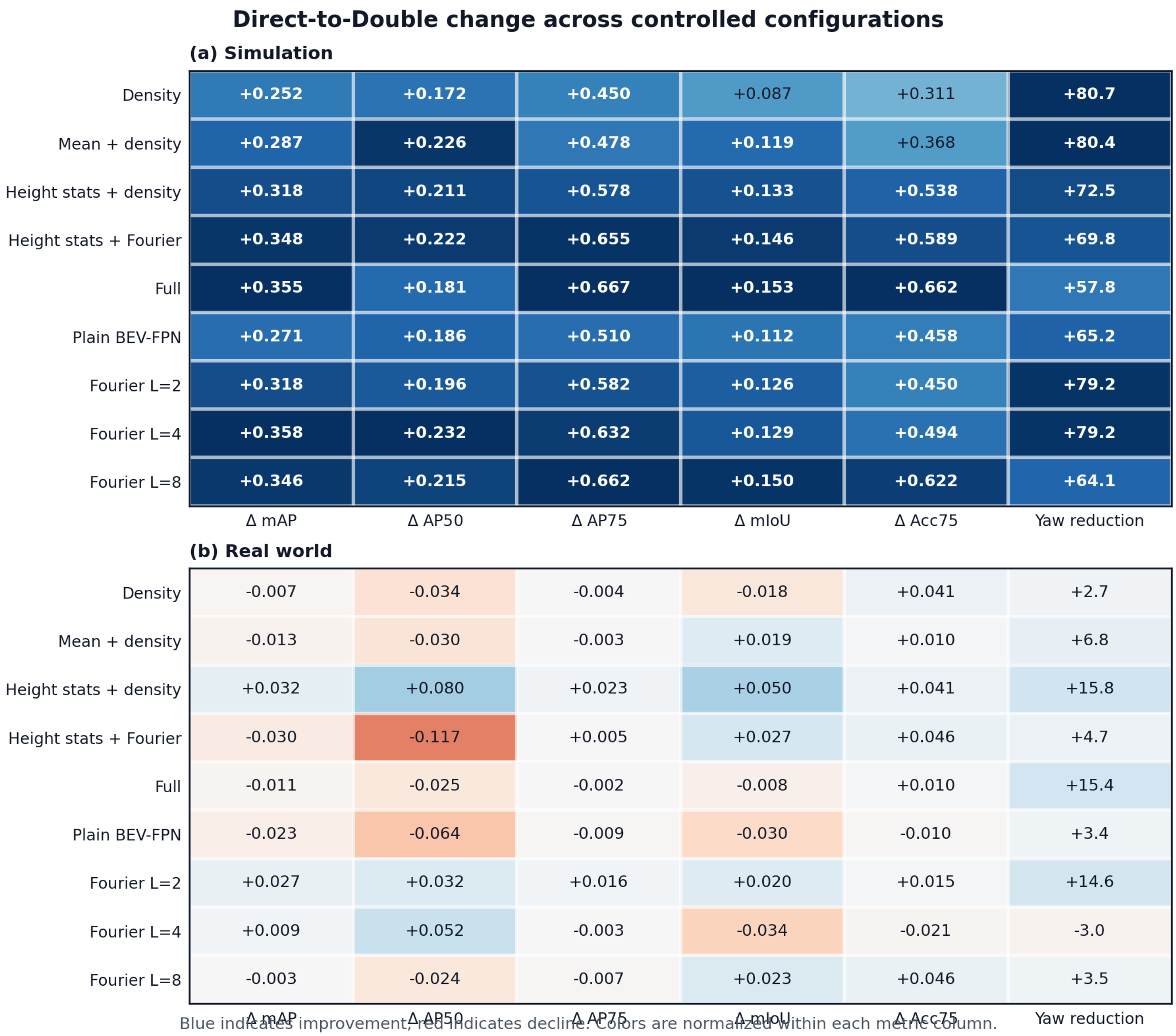

Figure 13: Complete paired changes from direct-angle to double-angle orientation regression across the controlled configurations. **a**, Simulated indoor dataset. **b**, Real-world robot dataset. For mAP, AP@0.50, AP@0.75, mean oriented IoU, and accuracy at IoU $0.75$, each cell reports the double-angle result minus the corresponding direct-angle result. The final column reports the reduction in median periodic yaw error, calculated as the direct-angle error minus the double-angle error. Positive values therefore indicate improvement in every column. Cell colours are normalized independently within each metric column to reveal configuration-level variation, while the annotations report the original, unnormalized changes.